\documentclass[pdflatex,sn-nature]{sn-jnl}

\usepackage{graphicx}%
\usepackage{multirow}%
\usepackage{amsmath,amssymb,amsfonts}%
\usepackage{amsthm}%
\usepackage{textgreek}
\usepackage[title]{appendix}%
\usepackage{textcomp}%
\usepackage{manyfoot}%
\usepackage{booktabs}%
\usepackage{algorithm}%
\usepackage{algorithmicx}%
\usepackage{algpseudocode}%
\usepackage{listings}%
\usepackage{lmodern}
\usepackage[table]{xcolor}
\usepackage{xr}
\usepackage[most]{tcolorbox}
\usepackage[version=4]{mhchem}

\tcbset{
    promptbox/.style={
        colback=white,                  
        colframe=magenta!20!white,      
        opacityback=0.9,                
        opacityframe=0.9,               
        fonttitle=\bfseries,
        coltitle=black,
        boxrule=0.8pt,
        arc=4pt,
        left=1em,
        right=1em,
        top=0.5em,
        bottom=0.5em,
        enhanced,
    }
}

\begin{document}

\title[Article Title]{Chemical reasoning in LLMs unlocks strategy-aware synthesis planning and reaction mechanism elucidation}


\author*[1,2]{\fnm{Andres} \sur{M Bran}}\email{andres.marulandabran@epfl.ch}
\author[1]{\fnm{Théo A.} \sur{Neukomm}}\email{theo.neukomm@epfl.ch}
\author[1]{\fnm{Daniel} \sur{Armstrong}}\email{daniel.armstrong@epfl.ch}
\author[1]{\fnm{Zlatko} \sur{Jončev}}\email{zlatko.joncev@epfl.ch}

\author*[1,2]{\fnm{Philippe} \sur{Schwaller}}\email{philippe.schwaller.epfl.ch}

\affil[1]{\orgname{\'Ecole Polytechnique F\'{e}d\'{e}rale de Lausanne (EPFL)}, \orgaddress{\street{Rte Cantonale}, \city{Lausanne}, \postcode{1015}, \country{Switzerland}}}

\affil[2]{\orgname{National Centre of Competence in Research (NCCR) Catalysis}, \orgaddress{\country{Switzerland}}}

\abstract{While automated chemical tools excel at specific tasks, they have struggled to capture the strategic thinking that characterizes expert chemical reasoning. Here we demonstrate that large language models (LLMs) can serve as powerful tools enabling chemical analysis. When integrated with traditional search algorithms, they enable a new approach to computer-aided synthesis that mirrors human expert thinking. Rather than using LLMs to directly manipulate chemical structures, we leverage their ability to evaluate chemical strategies and guide search algorithms toward chemically meaningful solutions. We demonstrate this paradigm through two fundamental challenges: strategy-aware retrosynthetic planning and mechanism elucidation. In retrosynthetic planning, our system allows chemists to specify desired synthetic strategies in natural language -- from protecting group strategies to global feasibility assessment -- and uses traditional or LLM-guided Monte Carlo Tree Search to find routes that satisfy these constraints. In mechanism elucidation, LLMs guide the search for plausible reaction mechanisms by combining chemical principles with systematic exploration. This approach shows strong performance across diverse chemical tasks, with newer and larger models demonstrating increasingly sophisticated chemical reasoning. Our approach establishes a new paradigm for computer-aided chemistry that combines the strategic understanding of LLMs with the precision of traditional chemical tools, opening possibilities for more intuitive and powerful chemical automation systems.}




\maketitle

\section{Main}
The automation of chemical reasoning has been a long-standing goal in many areas of chemistry, promising to accelerate drug discovery, retrosynthesis\citep{corey1969computer, schneider2018,guo2023can} and our understanding of chemical reactivity \citep{cheng2015computational}. Traditional machine learning and computational approaches have focused on specialized algorithms for specific tasks -- predicting properties \citep{fortunato2020data}, planning syntheses \citep{corey1969computer, mikulak2020computational}, or proposing reaction mechanisms \citep{zimmerman2013automated, Bradshaw2019a, Chen2009, kayala2012reactionpredictor, fooshee2018deep, tavakoli2021quantum, tavakoli2024, joung2025electron}. While successful in narrow domains, these systems lack the flexible reasoning and strategic multi-step thinking that characterize expert chemical problem-solving \citep{schwaller2019molecular, genheden2020aizynthfinder, coley2018}. A synthetic chemist planning the synthesis of a complex molecule, for instance, must simultaneously consider multiple strategic factors: which rings to form first, when to install sensitive functional groups and protective groups,  and how to leverage available starting materials \citep{corey1967general}. Additionally, this logical deconstruction of molecules must be supported by mechanistic reasoning which requires extrapolation of chemical principles of elementary steps and reactive intermediates, to newly proposed molecules, usually not observed before \citep{cheng2015computational,fey2022computational}.

The exponential growth in capabilities of Large Language Models (LLMs) has sparked a revolution across scientific disciplines \citep{phan2025humanity,mirza2024large,ruan2024automatic,guo2023can}, with applications spanning from automated literature analysis to hypothesis generation \citep{Kumbhar2025HypothesisGF,Cohrs2024LargeLM,Zimmermann2024ReflectionsFT}. These models have demonstrated an unprecedented ability to understand and reason about chemical concepts - from individual functional groups to complete synthetic pathways \citep{guo2023can,chen2311chemist,qian2023can}. Most remarkably, they exhibit reasoning patterns that mirror human chemical intuition rather than traditional computational approaches \citep{guo2025deepseek,mirza2024large}, showing particular promise in analyzing strategic elements of synthesis such as protecting group patterns and ring construction timing \citep{corey1967general}. However, a fundamental limitation persists: while LLMs excel at analyzing chemical concepts and strategies, they struggle to generate valid chemical representations, particularly SMILES strings \citep{jang2024llmsgeneratediversemolecules,silly_things_walters,edwards2022translation}, limiting their direct application in critical chemical tasks.

Here, we present a paradigm shift in how LLMs can advance chemical science: rather than attempting to generate chemical structures directly, we position these models as sophisticated reasoning engines that guide traditional search algorithms toward chemically meaningful solutions. This approach combines LLMs' ability to understand and evaluate complex chemical strategies with the systematic exploration capabilities of established search methods. Through systematic evaluation, we first demonstrate that LLMs can effectively analyze chemical entities and strategic patterns across multiple scales. We then show how these reasoning capabilities can be exploited in two challenging applications: strategy-aware retrosynthetic planning and mechanism elucidation.

Retrosynthetic planning ---the process of systematically breaking down complex target molecules into simpler, commercially available starting materials--- represents one of the most crucial and intellectually demanding tasks in organic chemistry \citep{corey1969computer,corey1985computer}. Current computational methods employ sophisticated search algorithms guided by either carefully designed or learned heuristics \citep{browne2012survey,Segler2017neural,chen2020retro}. While these systems excel at finding routes that end in commercially available materials \citep{genheden_bjerrum_2022,modelsmatter}, they often struggle to incorporate strategic considerations commonly used by expert chemists, such as optimal timing for ring construction, protecting group introduction or incompatible transformations \citep{corey1967general}. Recent specialized systems have addressed specific constraints like starting material availability \citep{armstrong2024tango,yu2024double} and bond preservation \citep{thakkar2023unbiasing,westerlund2024constrained}, but a framework for arbitrary strategy-aware synthesis planning has remained elusive. Our approach allows chemists to specify strategic requirements in natural language, with LLM-guided search identifying synthetic routes that satisfy these complex constraints. We demonstrate these capabilities on a challenging benchmark, along with a case study targeting global feasibility analysis and ranking of synthetic routes from multiple sources, including experimentally validated routes.

\begin{figure}[ht!]
    \centering
    \includegraphics[width=1.0\linewidth]{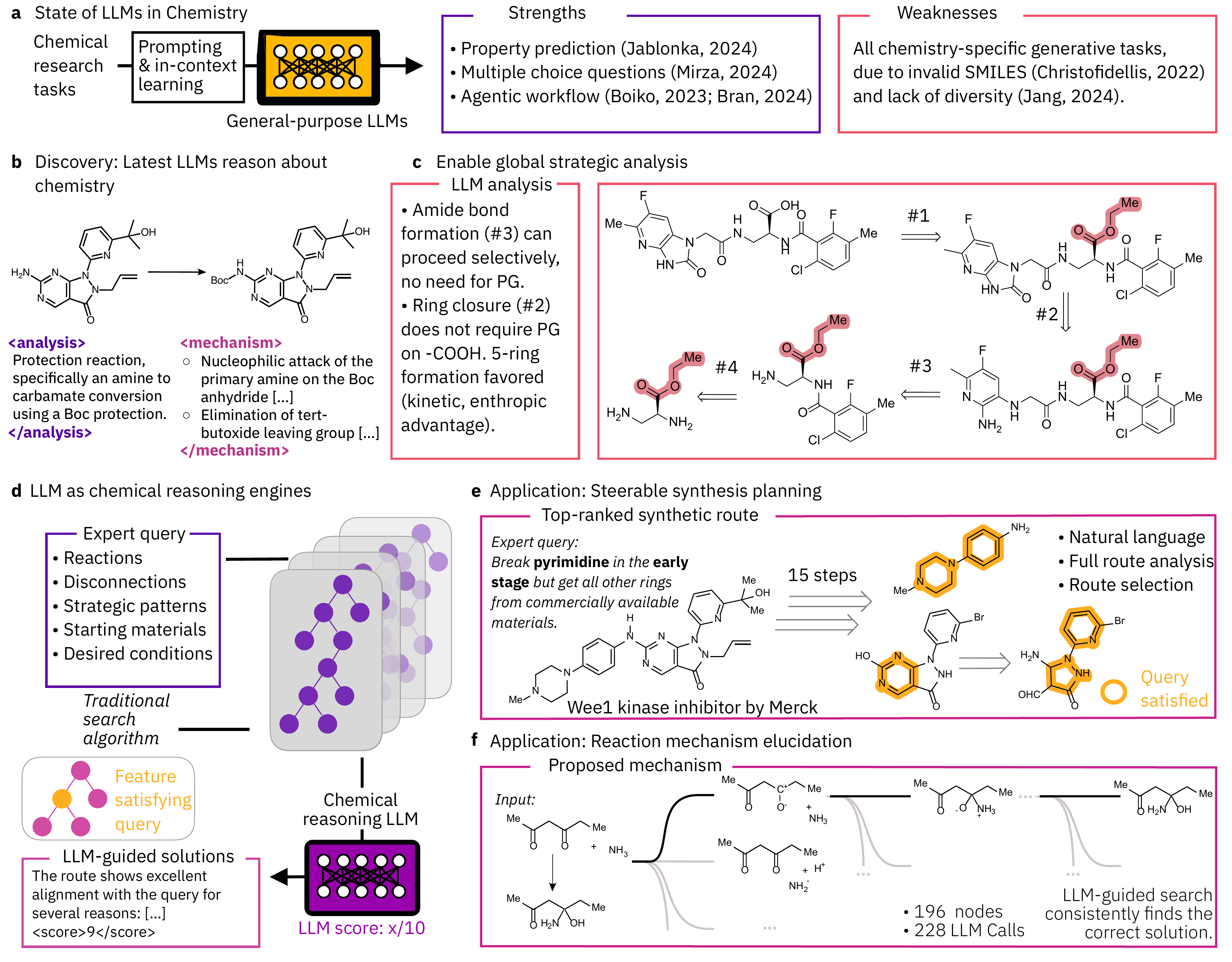}
    \caption{\textbf{LLMs as chemical reasoning engines for synthesis planning and mechanism elucidation.} a) Current state of LLMs in chemistry, highlighting strengths in property prediction, multiple choice questions, and agentic workflows, alongside limitations in structure generation tasks. b) LLMs demonstrate sophisticated chemical reasoning capabilities, providing detailed analyses of reaction mechanisms and functional group transformations. \textbf{c)} Analytical capabilities go beyond single reactions, being able to analyze a route in global terms. LLMs are capable of detecting when protecting groups are unnecessarily proposed as part of a synthetic route. \textbf{d)} Our key insight: positioning LLMs as strategic evaluators within chemical search frameworks. Rather than generating structures directly, LLMs guide traditional search algorithms toward chemically meaningful solutions. \textbf{e)} Application to synthesis planning: LLMs assess candidate routes based on guidance queries specifying strategic requirements (e.g., "break pyrimidine in the early stage"). This yields strategically relevant synthetic pathways with detailed rationales for synthetic choices. \textbf{f)} Application to mechanism elucidation: LLMs guide search through possible reaction mechanisms by evaluating the plausibility of elementary electron-pushing steps. The system efficiently identifies correct mechanistic pathways while providing chemically meaningful justifications. This approach combines the strategic understanding of LLMs with the precision of traditional chemical search algorithms.}
    \label{fig:overview}
\end{figure}

Similarly, mechanism elucidation ---understanding the step-by-step electron movements that transform reactants into products--- is fundamental to both chemical understanding and reaction optimization \citep{cheng2015computational,fey2022computational, murphy2018greenibuprofen}. The power of mechanistic understanding lies not only in explaining individual reactions but in their potential to generalize patterns of chemical reactivity that can be applied to previously unseen molecules \citep{clayden2012organic}. Existing computational methods can enumerate possible reaction paths \citep{zimmerman2013automated,zhao2021simultaneously} but often lack the chemical intuition needed to identify plausible mechanisms. While specialized approaches combining quantum calculations with search algorithms show promise \citep{kayala2012reactionpredictor}, they struggle to scale for complex systems and rely on predefined templates or atom-mapping, limiting their applicability. In our approach, basic electron-pushing steps are evaluated using LLMs' understanding of chemical principles in the context of a search algorithm, allowing the LLM to guide the search towards reasonable mechanisms, while potentially considering diverse forms of experimental evidence and practical constraints. Our results demonstrate that LLMs can effectively guide search processes and select optimal solutions, while providing chemically meaningful rationales for their decisions. Furthermore, we provide insights into how different models' capabilities ---related to pretraining \citep{kaplan2020scaling}, post-training \citep{dubey2024llama, ouyang2022training, guo2025deepseek} and inference-time scaling \citep{snell2024scaling,guo2025deepseek}--- affect solution quality, establishing crucial practical considerations for deploying such systems. This integration represents a significant step toward computational chemistry systems that can reason strategically about complex synthesis challenges while maintaining the precision of traditional computational tools.

\section{Results}


The approach presented here leverages LLMs as chemical reasoning engines that guide traditional search algorithms through complex chemical spaces. Here, a rule-based process is in charge of generating and proposing intermediate states, while the LLM serves as a judge to select among possible solution paths. The following results and our analyses show that current LLMs are capable of detailed analysis of relevant chemical objects like molecules, reactions, and reactive intermediates. Furthermore, we demonstrate how these capabilities can be leveraged in novel and useful manners for scientific discovery in synthetic chemistry through two challenging applications: prompt-guided and strategy-aware retrosynthetic planning, where natural language queries guide the search for synthetic routes with specific properties, and mechanism elucidation, where the goal is to identify plausible reaction mechanisms by evaluating candidate electron-pushing steps.

\subsection{Strategy-aware Synthesis Planning}
\label{sec:benchmark}

Retrosynthetic planning represents one of the most challenging tasks in organic chemistry, requiring both deep chemical knowledge and strategic thinking. While computational approaches have successfully automated the search through spaces of reactions, they typically struggle to incorporate the strategic elements that enable synthesis. The complexity of the task arises from multiple challenges across multiple levels and dimensions. At the reaction level, chemists must select transformations that are high-yielding, selective, and technically feasible. At a more global level, long-range considerations make the task almost an art: early synthetic decisions constrain later structural possibilities, protecting groups and functional group interconversions unlock otherwise impossible transformations, and these interconnected choices create intricate decision trees spanning entire synthetic routes. Expert chemists develop sophisticated heuristics for navigating these strategic elements. However, translating this expertise into computational systems has remained challenging.

Here, we extend traditional computer-assisted retrosynthetic search by incorporating natural language specifications of desired synthetic strategies. Given a molecular target and a description of desired route characteristics (e.g. "construct the pyrimidine ring in early stages"), the system must identify synthetic pathways that satisfy these strategic constraints, see Figure \ref{fig:overview}e. Such a task requires not only correct understanding of molecular and reaction representations but also an ability to connect chemical theory with practical experimental considerations and modern synthetic methods, depending on the query.

Considering the robust analytical capabilities demonstrated by LLMs (see Figure \ref{fig:overview}a-c, we developed Synthegy: a framework that combines these analytical capabilities with traditional synthesis planning software (Methods \ref{met:steerable-synth}). For evaluation, we created a benchmark consisting of pairs of molecular targets and prompts, along with scoring scripts that assess route-to-prompt alignment in a tailored manner (Methods \ref{met:steerable-bench}). These prompts specify features of the target synthetic route, and range from simple reaction preferences to complex strategic requirements, allowing us to assess the chemical reasoning capabilities of different LLMs and their ability to effectively guide synthesis planning, see SI-\ref{si:steering-prompts}.

\begin{figure}[ht!]
    \centering
    \includegraphics[width=0.90\linewidth]{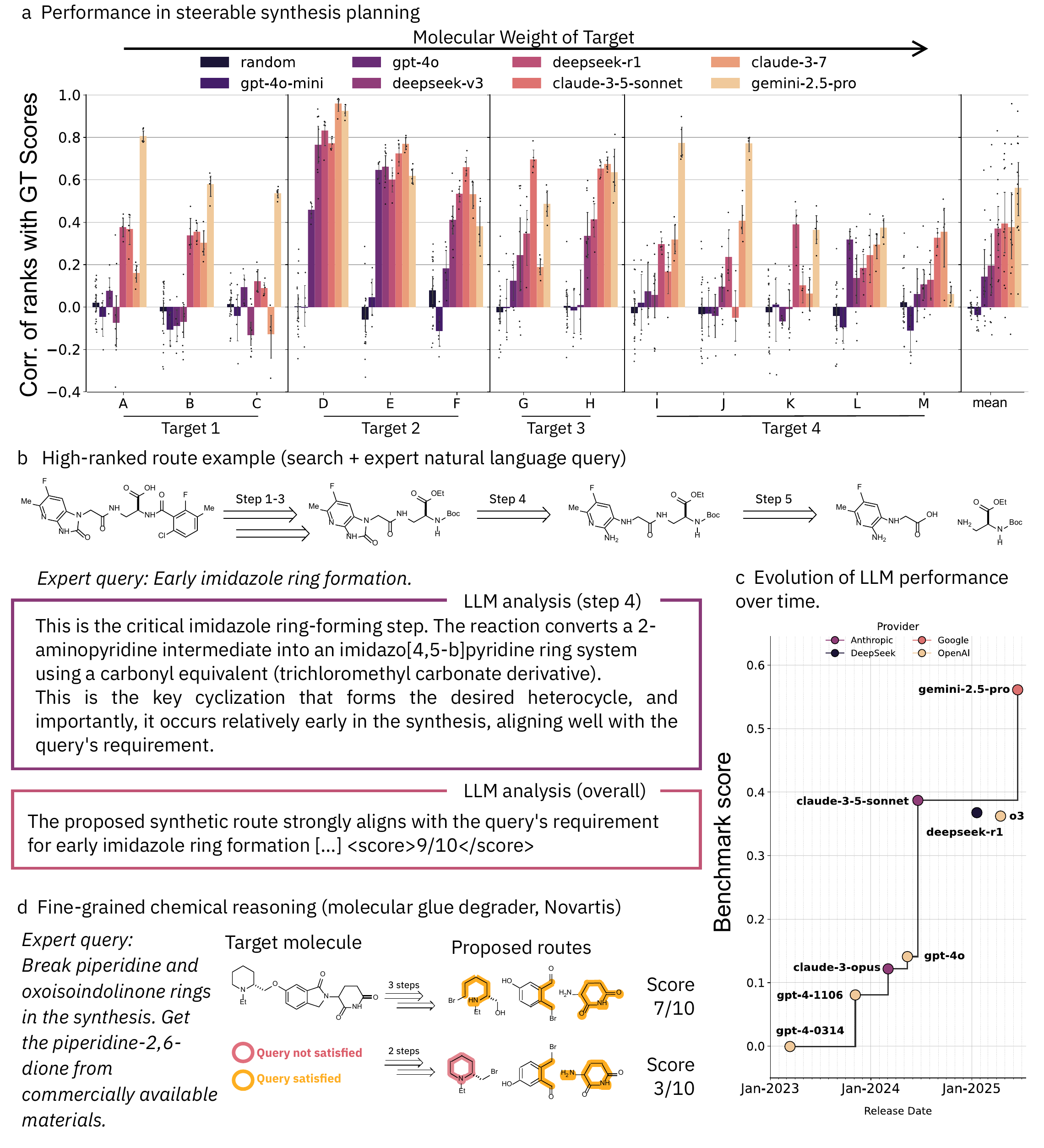}
    \caption{Performance of the system for strategy-aware synthesis planning presented in this work. \textbf{a} Performance of multiple LLMs of different sizes and providers across all the tasks in the benchmark. The tasks are grouped by synthetic target, and each column specifies a prompt as specified in SI-\ref{si:steering-prompts}. The y axis displays the correlation between LLM-produced scores and those computed as specified in the benchmark. Each data point represents a separate execution of the benchmark, with the number of repetitions (n) varying across LLMs based on execution time and resource constraints as specified in the legend. Error bars denote confidence intervals (95\%). \textbf{b} Example of an LLM's analysis of a synthetic route, where it provides a justification why a specific route received a high score. The example illustrates that the LLM analyses each reaction (exemplified with step 4), and then provides an overall analysis where it highlights the alignment with the user's query. \textbf{c} Average performance of multiple LLMs over time. Shown are only models that are the state of the art for their time, or close to it. The plot illustrates the rapidly evolving advancements in capability. \textbf{d} Illustrates the task of strategy-aware synthesis planning: a user specifies a target molecule along with a query in natural language, which specifies desired features in the route. The proposed solutions are given together with scores that signify their alignment with the query given by the user.}
    \label{fig:results-synth}
\end{figure}

Results in Figure \ref{fig:results-synth}a show that current commercial LLMs can already perform advanced reasoning about synthetic routes, successfully evaluating both specific reactions and global strategic features. Large, state-of-the-art models like Gemini-2.5-pro achieve the highest scores. This model achieves such performance by systematically analyzing each reaction in the synthetic sequence, while keeping track of the overall synthetic context, then correlating this with the given prompt (Figure \ref{fig:results-synth}b). Our results additionally show that the newer generation of models bring substantial improvements over older generations, with the newest Gemini-2.5-pro performing much better on the harder tasks (Target 4), where no other model could succeed. Figure \ref{fig:results-synth}c further highlights the rapid progress that is being achieved thanks to advancements in LLM capability. The plot shows that the kind of intelligence required to tackle the challenge of strategy-aware synthesis planning was entirely missing until around June 2024 with the release of Claude-3.5-Sonnet, which achieves an important improvement in capability relative to any other model at the time. More recently, gemini-2.5-pro established a new state of the art by improving over 50\% relative to the next best model. Notably, an open model like DeepSeek-r1 achieves high performances as well, positioning it as a strong open and local alternative to closed providers.

Performance scales strongly with model size. Smaller models' performance is indistinguishable from random choice even for short routes (see SI-\ref{app:more-performances}), suggesting that the chemical reasoning capabilities required for this task emerge only at larger scales. This suggests either a minimum threshold of model complexity required for meaningful chemical analysis, or a limitation in the multi-task performance of smaller models, suggesting that fine-tuning and other post-processing techniques might be needed to leverage small models on these tasks. Models from other providers (OpenAI \cite{openai_gpt4}, DeepSeek \cite{deepseek_r1}) show comparable performance patterns to Claude, with variations potentially attributable to differences in tokenization and prompting strategies rather than fundamental capability gaps. 

A key limitation is observed across all models when tackling long synthetic sequences. For the most complex targets (e.g. Target 4, involving routes up to 50 reactions), all models struggle to distinguish and select aligned routes, with the exception of the latest Gemini-2.5-pro \cite{google_gemini}, which demonstrates unprecedented performance in our benchmark, while producing highly reliable and faithful reasoning traces that enable explainability of the results, see SI-\ref{app:some-gemini-outputs} for examples. Despite these impressive results, some erroneous outputs are still observed. Common failure modes observed across different LLMs include grouping reactions instead of analyzing them individually, which causes overlooking of important fine details, and failing to correctly position individual reactions in the context of a large route. Still, LLMs show overall positive correlations in these extreme cases, showing that some relevant features of routes are still recovered by LLMs, making them useful even for the most complex routes, suggesting their utility as strategic evaluators. 

These results demonstrate that current LLMs can effectively assess synthesis planning through analysis and evaluation of arbitrary strategic elements in synthetic routes. In the sections following, we further analyze and validate these findings against real-world chemistry and use-cases, where alternative methodologies can hardly be applied. In particular, we focus on global assessments of feasibility. Under the Synthegy framework, this corresponds to \textit{steering} towards --or selecting-- routes that are overall more feasible, or that are likely to lead to the product in overall high yields.

\subsection{Assessing Feasibility of Synthetic Routes}
\label{sec:feasibility}

The examples demonstrated in Section \ref{sec:benchmark} show the power of Synthegy for selecting synthesis solutions as described in a prompt. The examples shown so far are verifiable by tailored cheminformatics code, thus making their value mostly demonstrative and evaluative, due to the possibility of defining ground truth answers. To test the limits of LLM capability on Synthegy ---beyond what current alternative tools allow--- we now use the same framework to attempt to select \textit{more feasible} synthetic routes. Feasibility involves assessing several factors and their interplay: prediction of side-products, hypothesizing and evaluating reaction conditions, feasibility of specific reaction types, functional group incompatibilities, etc. Beyond that, the \textit{strategic feasibility} of a synthetic route also plays an important role in this regard, as unnecessary reactions can severely hinder overall yields and compromise scalability.

With the same framework as introduced in the previous section, we use the following prompt to find feasible synthetic routes for the targets 1 to 4, see SI-\ref{si:steering-prompts}. 

\begin{tcolorbox}[promptbox,title=Feasibility prompt]
Highly feasible synthesis with high overall yields, consider potential side reactions and byproducts. Also ensure no unnecessary reactions are performed.
\end{tcolorbox}

We furthermore evaluate the feasibility of routes obtained through 3 different retrosynthesis engines: AIZynth \cite{genheden2020aizynthfinder}, Reaxys \cite{reaxys}, and Synthia$^{\text{TM}}$ \cite{synthia} (see Methods \ref{met:feasibility}). To compare these results against experimentally-validated routes, we also evaluate the original routes for these molecules as disclosed in their respective publications \cite{sabat_design_2024,matheson_development_2018,zhu_activating_2024}. All the results presented here were computed using Gemini-2.5-pro \cite{google_gemini} as a backend LLM as it demonstrated the best performance in our evaluations (see Section \ref{sec:benchmark}).

For Reaxys and Synthia$^{\text{TM}}$, we used the default preset settings, with routes restricted to the top-25 pathways out of the 50 generated ones for the latter. Those retrosynthesis engines can be further customized to user requirements, which would potentially lead to better results. For instance, Synthia$^{\text{TM}}$ can be fine-tuned according to chemical preferences, such as using robust chemistry and avoiding gaseous reactants or metal catalysis. Promoting more common reaction classes could better align with our score, as common reactions are classified as more feasible in our approach.

\begin{figure}[ht!]
    \centering
    \includegraphics[width=0.99\linewidth]{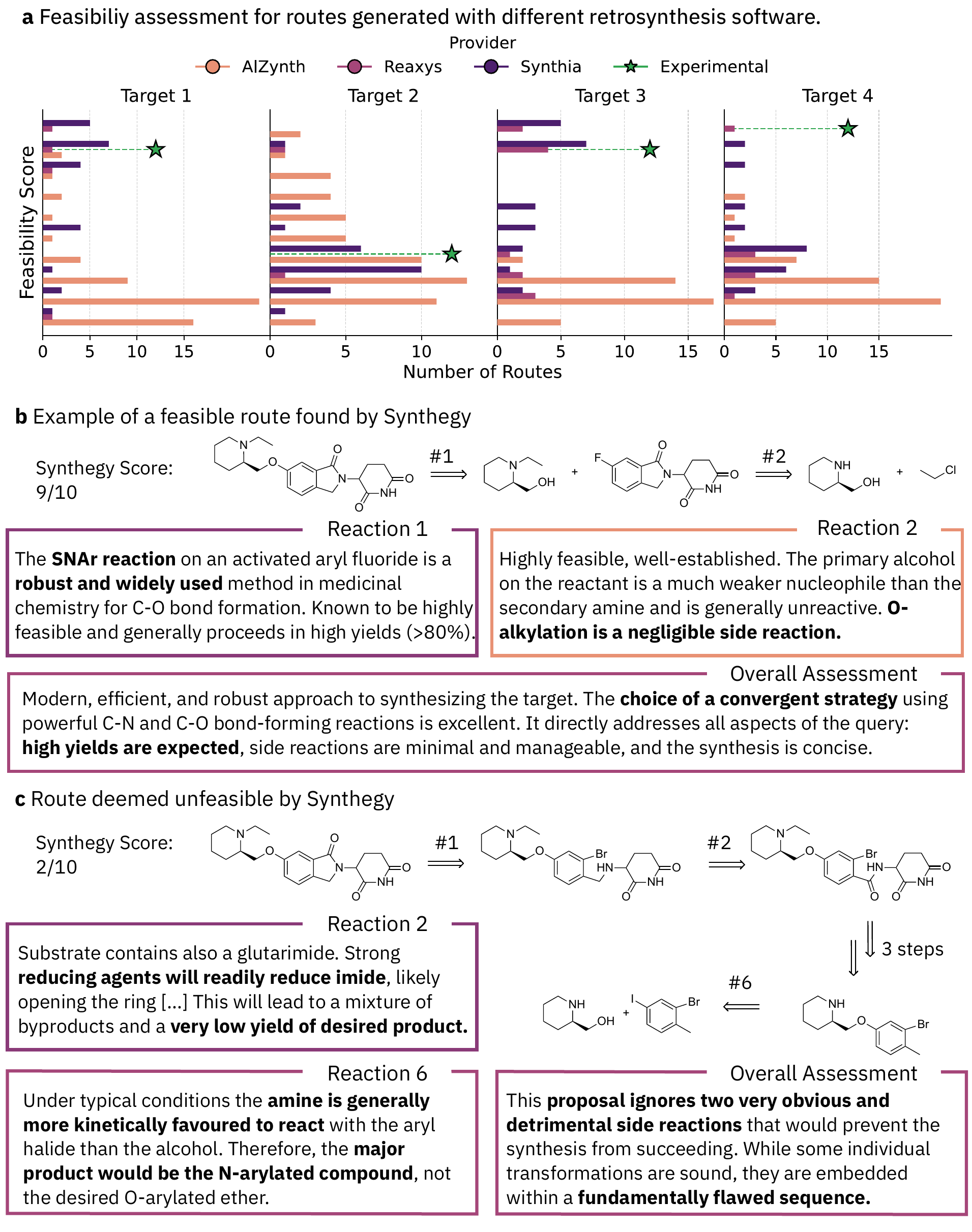}
    \caption{\textbf{Selecting highly feasible routes.} \textbf{a} Distribution of feasibility scores as determined by Synthegy, for the four targets in SI-\ref{si:steering-prompts}. Results have been computed for solutions coming from 3 different retrosynthesis engines and one experimentally validated route, for each target. \textbf{b} Example analysis of a feasible synthetic route. The model determines that the two reactions are feasible and the full plan is consistent. \textbf{c} Example of an unfeasible synthetic route. Synthegy correctly identifies 2 key flaws with the presented plan: namely an unfeasible last step, and an inefficient and illogical reaction sequence in preparation for a coupling reaction.}
    \label{fig:feasibility}
\end{figure}

The results in Figure \ref{fig:feasibility}a indicate a general bias towards generally infeasible routes in the case of AiZynth, with some small amount of these receiving high scores and thus being deemed feasible. The LLM justifies these low scores with several types of flaws, including ill-conceived protection strategies, reactions with low chances of success, and global strategic flaws such as unnecessary protection cycles and inappropriate ordering of reactions, see SI-\ref{app:why-low-score} for more details. Generally, the commercial tools perform the best as they tend to propose highly feasible solutions in most cases, and in all cases at least one highly feasible route. 

Evaluation of the original synthetic routes devised for these targets \cite{sabat_design_2024,matheson_development_2018,zhu_activating_2024} provide experimental validation of the power of Synthegy at assessing the feasibility of synthetic routes, as these routes were designed by expert chemists and actually executed in a lab, in enough quantities as to allow the evaluation of drug-relevant properties. As shown in Figure \ref{fig:feasibility}a, the experimentally realized routes obtain generally high scores, thus providing experimental evidence of the predictive power enabled by Synthegy. Interestingly, results for Target 2 show a general bias towards low scores, even for the experimentally evaluated route, however all tools provide at least one route scoring much higher than the experimentally validated one. Low scores for the experimental route for Target 2 may come from inefficiencies introduced as a result of batch synthesis for screening purposes, where a single route is engineered to synthesize multiple targets with a same scaffold, while varying R groups. 

Further inspection of the most feasible solutions reveals that Synthegy enables selection of highly elegant and feasible routes, avoiding protecting group cycles and selecting robust reactions, as shown in Figure \ref{fig:feasibility}b. Poorly conceived routes are rapidly discarded by Synthegy (see Figure \ref{fig:feasibility}c), citing redundant steps, high potential for side-products and low yields in critical steps, poor regioselectivity, among others, see SI-\ref{app:why-low-score}.

This experiment demonstrates that, given a large and diverse enough space of retrosynthetic solutions for a given target, Synthegy can identify and highlight those that most align with a given query. Such queries can range from simple reaction specifications or relative timing of scaffold formation, to complex feasibility assessments otherwise unfeasible.

\subsection{Mechanism Elucidation via LLM-Guided Search}

A reaction mechanism is a specification of why and how a given chemical transformation occurs, by means of a set of elementary steps \citep{clayden2012organic}. The power of mechanisms in chemistry lies not only in its explanatory power of a single reaction instance, but also in that the reach of an explanation may extend further than only that reaction; potentially explaining more observed reactions, and also even predicting potential unknown transformations \cite{murphy2018greenibuprofen}.

\begin{figure}[ht!]
    \centering
    \includegraphics[width=\linewidth]{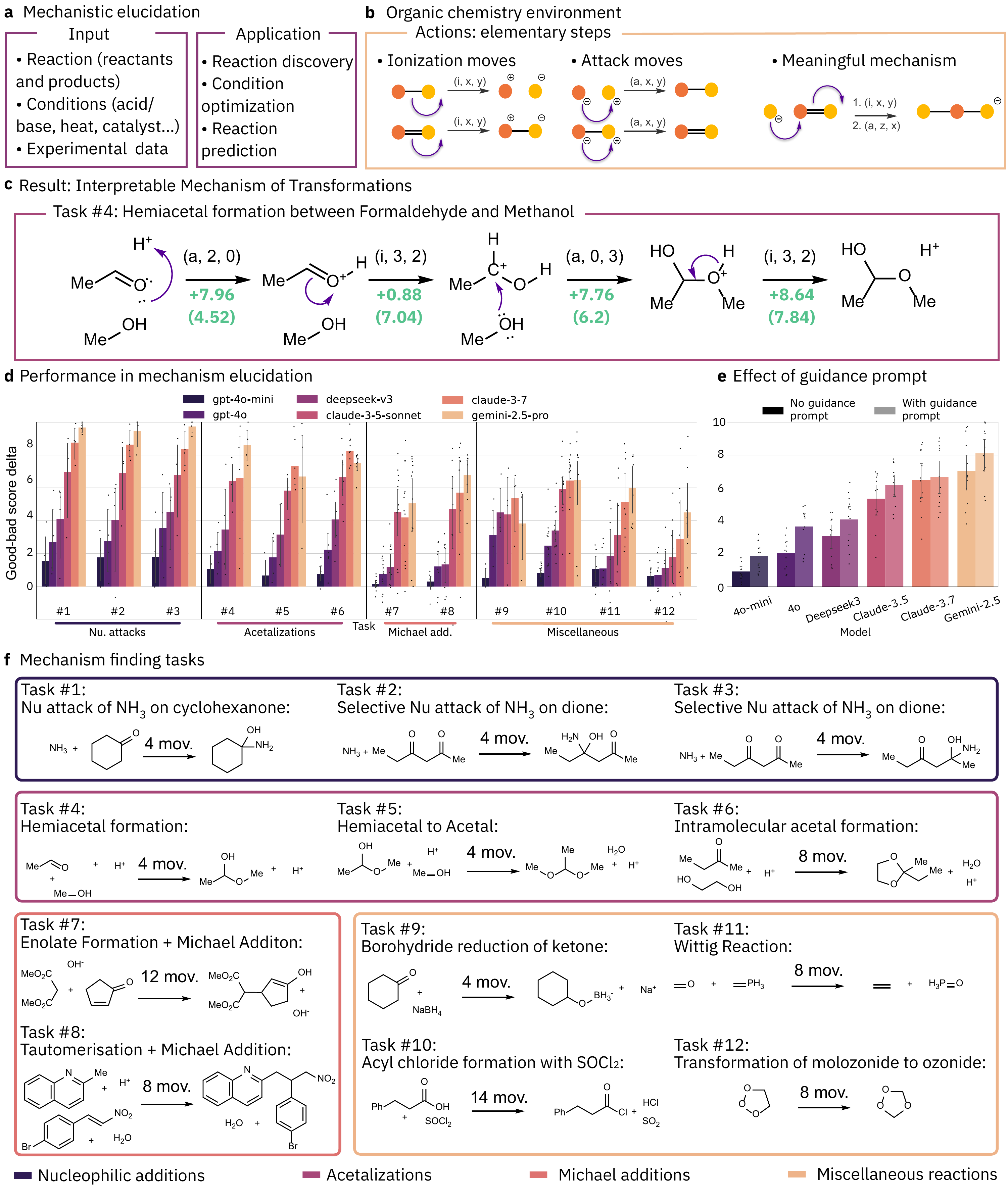}
    \caption{\textbf{Mechanisms elucidation} a) Requirements and impact of mechanistic elucidation. b) Example of actions in our mechanism framework displayed on example structures, with a post-processing interpretation. c) Example task broken down into moves compliant with our mechanism game framework. d) Model separation performance, averaged across the full mechanism in each case, with n=5 repetitions. e) Comparison of global performance with and without guidance prompts. f) All starting reactants and goal products of our 12 tasks, grouped by category.}
    \label{fig:mechanisms}
\end{figure}

We define a set of elementary steps that can be applied to any given molecular set (Methods \ref{sec:mechanism elementary steps} and Figure \ref{fig:mechanisms}b); these elementary steps can generally be applied to any bond/atom in any molecule, and thus serve as a fundamental basis for formulating mechanisms. The task is then to, given a chemical reaction, find a suitable sequence of such elementary steps that connects the reactants with the products. The search for the best path between reactants and products in chemistry has already been tackled using different ML paradigms, whether using generative models in rule-based environments \cite{Bradshaw2019a}, contrastive learning \cite{tavakoli2024}, reinforcement learning in 3D space \cite{barrett2024reinforcement} or flow-matching \cite{joung2025electron}. Here, we show that LLMs reasoning can be useful for this, even in complex cases. In our approach, an LLM analyzes a partially constructed solution and rates a proposed mechanistic step that continues such solution. For evaluation, we designed a benchmark comprising 12 diverse reactions (Figure \ref{fig:mechanisms}f) along with their mechanisms (SI-\ref{si:mechbench-rxns}). Performance is measured as the relative score gap between ground truth and alternative moves, averaged over all the steps in the mechanism (Methods \ref{met:eval-mmetric-mech}). A perfect scoring model would highly rate correct over incorrect moves, thus this metric correlates with performance in a real search setting by assessing the selectivity of the model at each step.

As shown in Figure \ref{fig:mechanisms}d, the best model evaluated achieves close to perfect performance on simple reactions like nucleophilic attacks, and poorer performance on more complex tasks like Michael additions on larger molecules and miscellaneous reactions with more complex mechanisms. Despite this drop in performance, the best models can still distinguish good moves over alternative incorrect ones, indicating good adaptability of their chemical knowledge into new situations. Smaller models like gpt-4o-mini generally perform badly, with poor performance even on the simplest of tasks, hinting at similar conclusions as in the previous sections: some complexity threshold might exist after which LLMs become smart enough for being useful at these tasks requiring strong chemical analysis.

\begin{small} 
\begin{table}[ht!]
    \centering
    \footnotesize
    \begin{tabular}{ccccccc}
    \toprule
    \multirow{3}{*}{\textbf{Task (\# steps)}} 
      & \textbf{4o Mini}
      & \textbf{4o}
      & \textbf{DS-3}
      & \textbf{Cd 3.5} 
      & \textbf{Cd 3.7} 
      & \textbf{Gm 2.5}\\ 
      \cmidrule{2-7}
      & \multicolumn{6}{c}{Random baseline: $17\%$}\\
\midrule

     1 (4) & $ 50 \pm 16$ & $ 55 \pm 19$ & $ 75 \pm  0$ & $ 90 \pm 12$ & $\mathbf{100 \pm  0}$ & $\mathbf{100 \pm  0}$\\
     2 (4) & $ 70 \pm 10$ & $ 60 \pm 12$ & $ 70 \pm 10$ & $ 95 \pm 10$ & $\mathbf{100 \pm  0}$ & $95 \pm  10$\\
     3 (4) & $ 55 \pm 10$ & $ 70 \pm 10$ & $ 80 \pm 10$ & $ 75 \pm  0$ & $\mathbf{100 \pm  0}$ & $\mathbf{100 \pm  0}$\\
\midrule
     4 (4) & $ 30 \pm 10$ & $ 50 \pm 16$ & $ 55 \pm 10$ & $ 75 \pm  0$ & $ 75 \pm  0$ &$ \mathbf{90 \pm 12}$\\
     5 (4) & $ 35 \pm 12$ & $ 50 \pm  0$ & $ 50 \pm 16$ & $ \mathbf{80 \pm 10}$ & $ \mathbf{80 \pm 10}$ & $ 70 \pm 19$\\
     6 (8) & $ 25 \pm  8$ & $ 52 \pm  5$ & $ 70 \pm 13$ & $ 80 \pm  6$ & $ \mathbf{85 \pm  5}$ & $ 78 \pm 39$\\
\midrule
     7 (12) & $ 20 \pm 10$ & $ 48 \pm  9$ & $ 20 \pm 15$ & $ 75 \pm  0$ & $ \mathbf{80 \pm  6}$ & $ 72 \pm 37$\\
     8 (8) & $  5 \pm  4$ & $ 18 \pm 10$ & $ 28 \pm  7$ & $ \mathbf{68 \pm  3}$ & $ \mathbf{68 \pm 10}$ & $ 67 \pm 34$\\
\midrule
     9 (4) & $ 30 \pm 10$ & $ 30 \pm 10$ & $ 60 \pm 25$ & $ 30 \pm 10$ & $ \mathbf{80 \pm 10}$ & $ 40 \pm 30$\\
     10 (14) & $ 24 \pm 10$ & $ 43 \pm  6$ & $ 36 \pm 12$ & $ 66 \pm  3$ & $ \mathbf{79 \pm  5}$ & $ 64 \pm 23$\\
     11 (8) & $ 32 \pm  6$ & $ 28 \pm  9$ & $ 35 \pm  5$ & $ 50 \pm  8$ & $ \mathbf{62 \pm  8}$ & $ 57 \pm 30$\\
     12 (8) & $ 22 \pm  9$ & $ 18 \pm  6$ & $ 18 \pm  6$ & $ 18 \pm  6$ & $ 42 \pm  6$ & $ \mathbf{55 \pm 19}$\\
\bottomrule
    \end{tabular}
    \caption{\textbf{Top-1 accuracy on mechanisms elucidation tasks} Percentage of ground truth moves classified as strict top-1 when scored with 5 alternatives at each step. Each row displays the results for each of the tasks on the left column. The number of moves involved in each task are specified in parentheses. The reported scores are percentage of correctly scored moves. Scores are reported for LLMs alone, without the help of a guidance prompt defining the mechanism. Each column displays the results for each LLM. The column names used stand for: 4o-mini: GPT-4o-mini, 4o: GPT-4o, DS-3: DeepSeek-3, Cd 3.5: Claude-3.5-Sonnet, Cd 3.7: Claude-3.7-Sonnet, Gm 2.5: Gemini-2.5-pro.}
    \label{tab: mech percentages top 1 alone}
\end{table}
\end{small}

A key advantage of search guided by an LLM is that it allows the specification of any arbitrary amount of information, parameters and instructions, through the LLM's text interface. Particularly relevant for this task, the input consists of the reaction (reactants and products), but could also contain the reaction conditions (solvent, temperature, concentrations, etc), any available experimental data (such as kinetic studies), among others. In this line of ideas, we experiment with text-guided search, where search is guided by an external text describing the sequence of steps in a hypothetical mechanism. The search here thus functions as a decoder from the input text+reaction, into a proper sequence of elementary steps. The source of this external text can be an expert human, or another LLM which, as is clear from the previous sections, can accurately analyze reactions and describe their mechanisms. Figure \ref{fig:mechanisms}e shows that adding such information is generally beneficial for performance, boosting the performances of poorly performing LLMs, as is the case of gpt-4o-mini and gpt-4o. Notably, even though Gemini-2.5-Pro's base performance is already higher than any other competitor, guidance prompts still provide additional and substantial improvements. These results show the potential for including multiple sources of information as input, hinting at the future possibility of automated mechanistic (hypothesis) generation and closed-loop refinement.

\section{Discussion}


In this work, we demonstrate that LLMs can serve as powerful tools enabling chemical analysis at arbitrary levels of abstraction. We present Synthegy: a framework to exploit these capabilities in scientifically relevant tasks in synthetic chemistry. Our approach combines traditional search algorithms with LLMs' sophisticated reasoning abilities, positioning the models as expert evaluators that can filter, rerank, or guide search toward solutions that are aligned with chemical principles as well as expert intent, across a wide rang of chemical scales. This integration helps overcome fundamental limitations in existing tools while enabling more intuitive interfaces for complex chemical tasks.

Our results first establish that current LLMs possess remarkable capabilities for detailed chemical reasoning, accurate analysis and evaluation of chemical objects at multiple scales, from functional groups in molecules, to single reactions, to strategic patterns in full synthetic routes. We leverage these capabilities through a framework where candidate solutions are generated by traditionally used computational environments, while LLMs analyze proposed solutions to assess their validity in relation to the specified target solution. Furthermore, these capabilities can be leveraged to directly guide search algorithms towards desirable and chemically-sensible solutions.

We showcase the practical utility of this approach through two challenging applications: strategy-aware synthesis planning and mechanism elucidation. In synthesis planning, our system enables natural language specification of strategic constraints, allowing chemists to select solutions from computer assisted synthesis planning algorithms, towards routes that are more aligned with their intent. Such intent can range from specifications of the reaction sequences used or relative timings of scaffold formation, to full assessments of general feasibility, enabling a novel way of selecting promising and experimentally feasible synthetic routes. We demonstrate this through an experimental study where Synthegy successfully selects the experimentally validated route as highly-feasible, from a non-exhaustive pool of candidate solutions proposed by several standard retrosynthesis tools.

We furthermore evaluate the system on experimentally validated routes for complex targets like Strychnine, and show how the system correctly identifies and evaluates key strategic elements, not only providing key insights into the analyzed solutions but also further validating the practical potential of Synthegy to select solutions with specific properties, even for complex molecular scaffolds and strategic requirements.

Similarly for mechanism elucidation, we show that a similar approach can successfully guide search towards chemically feasible mechanisms, starting only from a set of simple and general elementary steps, making our approach generally applicable to a large space of organic reactions. We note that any arbitrary specification, guidance, or initial condition can be encoded as input to the system, thanks to the flexibility of the LLMs' text interface. We show the potential of this idea by experimenting with "guidance prompts" in the input, which resemble intuitions of expert chemists about how the mechanism of a given reaction should look like. These considerations can be extended to include feedback from experiments, that iteratively refine mechanistic hypotheses.

However, some limitations remain. LLMs exhibited a number of failure modes, more prominently eventual misunderstandings of the format of the input data, which drives them to analyze reactions in an inverse sense and thus leading them to deem routes unfeasible. Furthermore, some models display lazyness, which bias models' responses towards shorter and overly simplistic and optimistic responses. Such behavior can be traced back to pre- and post-training techniques used to train these models. These limitations can be tackled by improved prompting techniques, task-specific fine-tuning, among others. Even with these limitations, we show that the current models are very strong baselines, highlighting the potential for future improvements.
Moreover, we show that success in the current implementation is very much dependent on the quality and diversity of synthetic routes generated by the backend retrosynthesis systems. Further research needs to go into the application of these strong capabilities for directly guiding search methods towards solutions that align with prompts and chemical principles, overcoming the inherent limitations of predefined sets of solutions.

Looking forward, our framework opens new possibilities for computer-aided chemistry systems that better align with human chemical intuition while maintaining computational precision. The ability to incorporate diverse forms of chemical knowledge ---from initial conditions, to practical constraints, experimental data, or even expert intent--- through natural language interfaces could make advanced computational tools more accessible to practicing chemists while enabling new forms of human-AI collaboration in chemical research.

\newpage
\section{Methods}

\subsection{Large Language Models}
\label{met:llms}

Throughout this work we used several LLMs from different providers, all through litellm \cite{litellm} as it provides a unified interface to multiple provider's APIs. A temperature of 0.1 has been used across all LLMs. The specific models used throughout this work include: OpenAI models \cite{openai_gpt4} (gpt-4o, gpt-4o-mini, gpt-4-turbo, o3, o4-mini), Anthropic Claude models \cite{anthropic_claude} (claude-3-sonnet-20240229, claude-3-opus-20240229, claude-3-5-sonnet-20241022, claude-3-7-sonnet-20250219, claude-opus-4-20250514, claude-4-sonnet-20250522, claude-2.1) , Meta LLaMA models \cite{meta_llama} (llama-3.3-70b-instruct, llama-3.3-8b-instruct, llama-4-scout, llama-4-maverick, llama-3.1-405b-instruct), DeepSeek models \cite{deepseek_r1} (deepseek-r1-0528, deepseek-chat-v3-0324, deepseek-r1-distill variants), Google Gemini models \cite{google_gemini} (gemini-2.5-pro-preview, gemini-flash-1.5, gemini-pro-1.5, gemini-2.0-flash-001, gemini-2.5-flash-preview), and Qwen \cite{qwen2.5} (qwen3-32b).

\subsection{Strategy-aware synthesis planning}
\label{met:steerable-synth}

In this work we present a framework for strategy-aware synthesis planning, which integrates LLMs as evaluators within a traditional retrosynthetic search algorithm. The process begins with a target molecule and a natural language query specifying desired strategic features of the synthetic route. We employed the AiZynthfinder software \citep{genheden2020aizynthfinder} as the underlying retrosynthesis engine to generate a diverse set of potential synthetic routes for a given target molecule. For each candidate route generated by the underlying retrosynthesis software, we constructed a detailed textual representation of the route, including SMILES strings of all intermediates and reactants, and the sequence of transformations applied at each step. This textual representation, along with the user-provided natural language query, is then passed to the LLM. 
The LLM is prompted to analyze the synthetic route and evaluate its alignment with the strategic requirements specified in the query. The LLM's output is a score reflecting the degree to which the route satisfies the user's query, along with a textual rationale justifying its assessment. This score is then used to rank and filter the candidate routes generated by AiZynthfinder, prioritizing routes that are deemed strategically relevant by the LLM. This process allows us to leverage the systematic exploration capabilities of AiZynth while incorporating the sophisticated chemical reasoning of LLMs to guide the search towards strategically desirable synthetic pathways.

\subsection{Feasibility assessment}
\label{met:feasibility}

To evaluate the capability of Synthegy to assess synthetic route feasibility, we collected synthetic routes from multiple sources and compared their LLM-assigned feasibility scores. Routes were obtained from three different retrosynthesis engines alongside experimentally validated routes from the literature.

We accessed synthetic routes through the following platforms:

\textbf{AiZynth:} Synthetic routes were obtained using the AiZynth retrosynthesis planning software \cite{genheden2020aizynthfinder} with default configuration parameters. The software was run locally using standard installation procedures and default model weights.

\textbf{Reaxys: }Synthetic routes were obtained through the Reaxys API using default parameters as specified by the provider. Access was granted through institutional licensing agreements. The API was queried using target molecule SMILES strings, and the top-ranked routes (by Reaxys scoring) were collected for each target.

\textbf{Synthia:} Routes were generated using Synthia-Lite through their web interface with default parameter settings. No specific parameter customization was performed, utilizing the platform's standard search algorithms and scoring functions. Routes were collected in SMILES format for subsequent analysis.

\textbf{Experimental Routes:} Literature routes for targets T1-T4 were manually extracted from their respective publications \cite{sabat_design_2024,matheson_development_2018,zhu_activating_2024} and converted to SMILES format following the same reaction representation standards used for computationally generated routes.

\paragraph{Feasibility Scoring:} All collected routes were evaluated using the Synthegy framework with gemini-2.5-pro as the backend LLM, selected based on its superior performance in our benchmark evaluation (Section \ref{sec:benchmark}). The feasibility assessment prompt (Section \ref{sec:feasibility}) was applied consistently across all route sources. Each route was scored independently, with the LLM providing both numerical scores (0-10 scale) and detailed rationales for feasibility assessments.

\subsection{Strategy-aware synthesis planning - Benchmark}
\label{met:steerable-bench}

To evaluate the performance of our strategy-aware synthesis planning framework, we created a benchmark consisting of pairs of molecular targets and prompts. We selected a diverse set of molecular targets that represent various levels of complexity and synthetic challenges. These targets include both well-studied molecules and novel compounds to ensure a comprehensive evaluation. For each molecular target, we crafted prompts that specified desired strategic elements in the synthetic route. These prompts ranged from simple reaction preferences to complex strategic requirements. The specific prompts used are given in the SI \ref{si:steering-prompts}.

As the system returns a set of routes, together with alignment scores, the primary evaluation metric reported in this work is correlation with ground truth scores. The scores are computed using scripts tailored for each target-prompt pair, and work by finding a specifically defined event in a synthetic route. For example, if the prompt requires "an early ring formation", then this is translated into a script that finds ring-forming reactions, together with the relative position of this reaction, if any, in the route. A score is thus calculated to account for the happening of the reaction, and it's relative position, in this case giving a higher score to routes where a ring-forming reaction occurs in the early stages (far from the final product), and progressively lower scores for routes when this happens in later stages. The minimum score is given to routes where the ring-forming reaction doesn't happen. The full scripts can be found in the github repository.

\subsection{Mechanism elementary steps}
\label{sec:mechanism elementary steps}
In order to create an exhaustive and computationally accessible state/action space, mechanisms have been broken down into their most elementary components. Rooted in the arrow-pushing formalism, the possible actions at each state (corresponding to a set of molecules) consist of two fundamental types: ionisation and attack moves. An ionisation move is defined as any bond decreasing its bond order by one, ionising at the same time on one of its terminal atoms. An attack move, on the other hand, is defined as any atom with a lone pair of electrons attacking any atom with an empty orbital, increasing the bond order between them by one, and charging them correspondingly. In both cases, a bond order of 0 corresponds to no bond existing between the two atoms. As minimalistic as this set of moves might seem, it has proven to be quite practical as a systematic way to enumerate and also to translate the majority of non-radical chemistry. However, a limit has been found regarding concerted moves. Even though such cases could often be detected and resolved via a post-processing process, cases like SN1 vs. SN2 (which explicitly require one to know if the electron moves are concerted or not) could not be resolved from the move sequence itself.

\subsection{Mechanism elucidation - Benchmark}

The mechanism elucidation benchmark was designed to include diverse cases of chemical reactions. It includes several reaction types as well as different scales of molecular weight of the products, assessing different levels of understanding of chemical principles, molecular and reaction representations, and ability to understand the rules proposed by the environment. In Figure \ref{fig:mechanisms}c and SI-\ref{si:mechbench-rxns}, it is shown how a mechanism is broken down into ionizations and attacks only.

\subsection{Mechanism elucidation - Metrics}
\label{met:eval-mmetric-mech}

For each elementary step of the 12 reactions, we computed five alternatives to the ground truth that are legal in our framework. At each step $n$ of a certain mechanism $m$, the LLM is prompted with the mechanistic rules defined here, along with the reactants and products of the reaction, as well as the history of ground-truth moves up to step $n-1$. The prompt also includes a possible next step --- one of the 6 in the benchmark --- and instructs the LLM to score such move in a scale from 0 to 10 according to alignment with its knowledge of chemical principles and the given reaction. The results presented here correspond to 5 independent runs on each step for reproducibility.
The main metric reported in Figure \ref{fig:mechanisms}d, is the score delta between the ground-truth option, and the mean score for the 5 alternative options, averaged across all steps for a given task.
Table \ref{tab: mech percentages top 1 alone} reports the percentage of moves for which the ground truth option was given as a strict top-1 (score strictly higher than any of the other possibilities).

\section*{Acknowledgments}

A.M.B. and P.S. acknowledge support from the NCCR Catalysis (grant number 225147), a National Centre of Competence in Research funded by the Swiss National Science Foundation. T.A.N. acknowleges support from Intel and Merck KGaA via the AWASES programme. D.A. and Z.J. acknowledge support by the Swiss National Science Foundation (SNSF) (grant number 214915). All the authors thank Merck MPS/Synthia and Reaxys for access to their retrosynthetic software and outputs for testing and evaluating the feasibility score.

\section*{Data and Code availability}
All benchmarks and datasets used, along with the code have been released at \url{https://github.com/schwallergroup/steer}. Access to the proprietary LLMs GPT-4, Claude, and DeepSeek can be obtained through each provider's API. All data generated in this work has been released at \url{https://figshare.com/s/bcf0d2c747e50b7f48d7}.

\bibstyle{naturemag}

\bibliography{iclr2025_conference}

\end{document}


\maketitle

\section{Chemical reaction reasoning capabilities}
\label{si:early-experiments}

Unless specified, all the experiments and results that follow were conducted with Claude-3.5-Sonnet as the LLM.

\subsection{LLM analytic capabilities on measurable tasks}

We examine how LLMs comprehend synthetic strategy through their ability to describe the starting materials used in a synthesis. Starting materials play a crucial role in determining synthetic strategy through providing pre-constructed structural motifs, functional groups and stereocenters to the synthesis pathway. An ability to parse, understand, and extract semantic value from starting materials can be viewed as a necessary pre-condition for LLMs to understand entire synthetic routes. To assess this, we ask an LLM to describe the synthetic route in relation to its starting materials. This is then embedded using OpenAI's embedding model "text-embedding-3-large", and a pairwise similarity matrix is constructed using cosine similarity. In parallel we extract the starting material's SMILES from the relevant routes, and compute a similarly-constructed pairwise matrix using size of intersection set as a similarity measure; this functions as the ground truth similarity matrix.

\begin{figure}[ht!]
    \centering
    \includegraphics[width=0.7\linewidth]{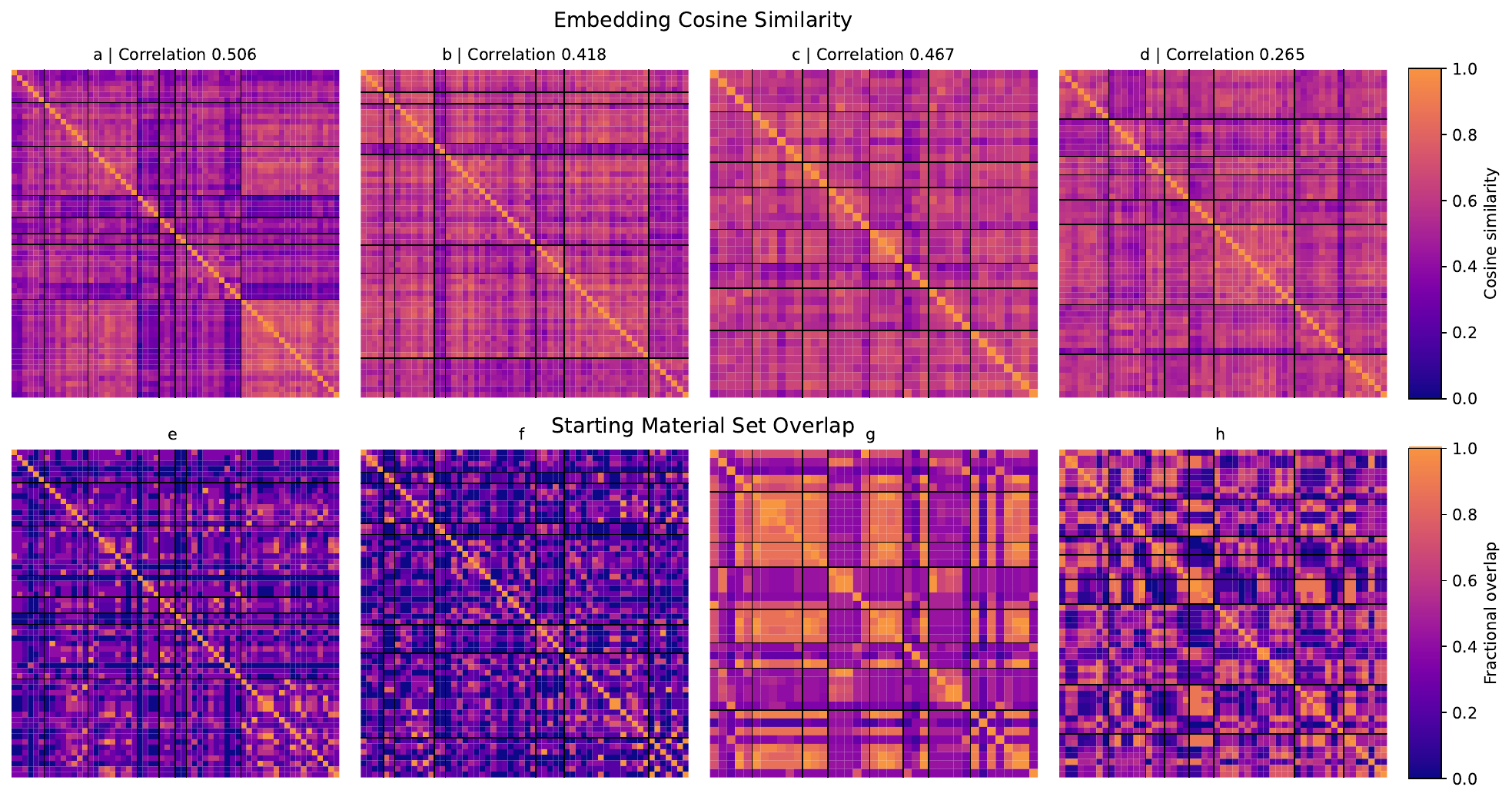}
    \caption{Comparison between different comparison methods between synthetic routes in terms of their starting materials. Top row shows the similarity computed as the cosine similarity between LLM descriptions of the routes in terms of their starting materials. The bottom row is a fractional overlap between the starting materials in each pair of routes. All plots are sorted according to clustering in the plots of the top row.}
    \label{app:sm_clusters_figure}
\end{figure}

The results in Figure \ref{app:sm_clusters_figure} show a relatively high correlation between the two route comparison methods, indicating that the LLMs can accurately describe synthetic routes, at least in terms of their starting materials. In addition, the figures displayed show similar patterns between the ground truth (bottom row) and the LLM-computed similarities (top row).

\begin{figure}[ht!]
    \centering
    \includegraphics[width=0.8\linewidth]{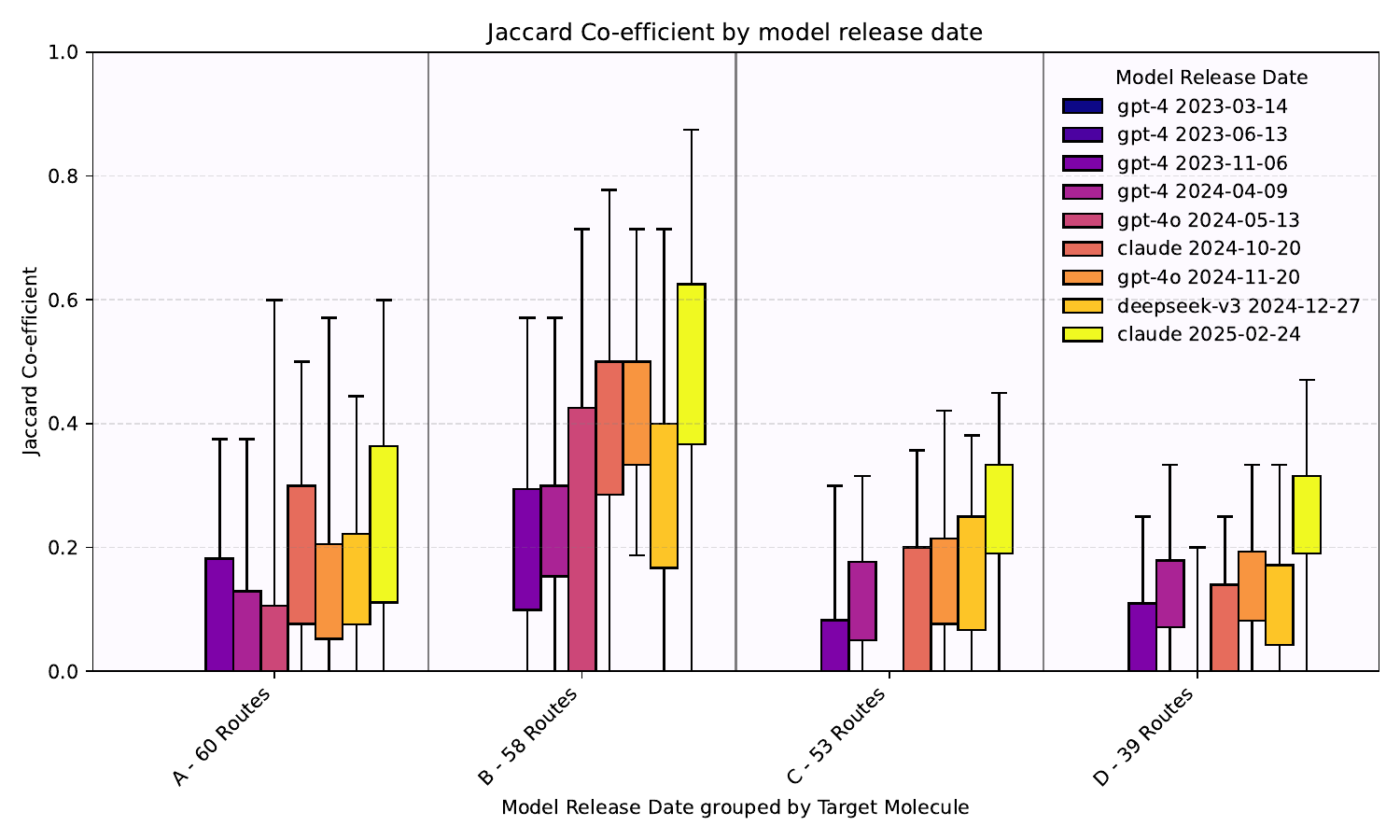}
    \caption{Jaccard similarity between ground truth functional groups in starting materials, and functional groups extracted with an LLM. The boxplots represent a distribution across multiple routes for a given target. The targets used are the same as shown in SI-\ref{si:steering-prompts}.}
    \label{app:fg_overlaps}
\end{figure}

For a slightly more advanced task, we construct an experimental setting where an LLM is tasked to extract all functional groups from the starting materials in a synthetic pathway. By doing so we directly assess whether LLMs grasp the chemical constraints that dictate the available reaction space and order of transformations. We use an in-house rule based system for functional group extraction to determine the ground truth set and measure  (treating a functional group as a token) the LLM's error using the Jaccard coefficient (Figure \ref{app:fg_overlaps}).

Claude-3.5-sonnet shows higher overall alignment with the rule-based ground truth, while GPT-4o-mini generates more functional groups but suffers lower precision. Both LLMs show significant variance in their output. Minor formatting and naming differences between the rule-based and LLM outputs obscure direct comparisons, likely understating the true accuracy of the LLMs. We leave an anecdote that LLMs manage to correctly extract additional ring-system functional groups which are not currently tagged by our rule-based approach.

\subsubsection{Strategic elements in synthetic routes}
 
Synthetic routes can be analyzed across multiple dimensions. Previous results show promise in synthetic route analysis at the level of single molecules, however strategy in synthetic routes is marked by a sequential, non-local analysis of the sequence of reactions \cite{Corey_Chelg_LogicofChemicalSynthesis}, where some steps are only performed in preparation for other steps down the line \cite{gajewska_algorithmic_2020}. An important pattern in synthesis is the use of protecting groups, which temporarily mask reactive functional groups to allow selective transformations \cite{schelhaas1996}. Correct understanding of protecting groups remains a major weakness of existing synthesis tools. These tools often propose either non-selective reactions that require protection, or conversely, include redundant protecting groups that add unnecessary steps \cite{hardy2022, latendresse2023}.

In a first experiment, we formulate the challenge as false positive detection, where LLMs identify routes where protecting groups are proposed unnecessarily. We instructed the LLM to analyze each synthetic route and tag them using a classification scheme:

\begin{itemize}
\item No protecting group: Routes that correctly do not use protecting groups
\item Protecting group needed not used: Routes requiring protection but lacking it
\item Protecting group not needed but used: Routes with unnecessary protecting groups
\item Protecting group needed and used: Routes that correctly use protecting groups
\end{itemize}
Routes tagged as either requiring but not containing , or containing unnecessary protecting groups were subsequently subjected to in-depth analysis to evaluate the model's reasoning capabilities.

\begin{figure}[ht!]
    \centering
    \includegraphics[width=\linewidth]{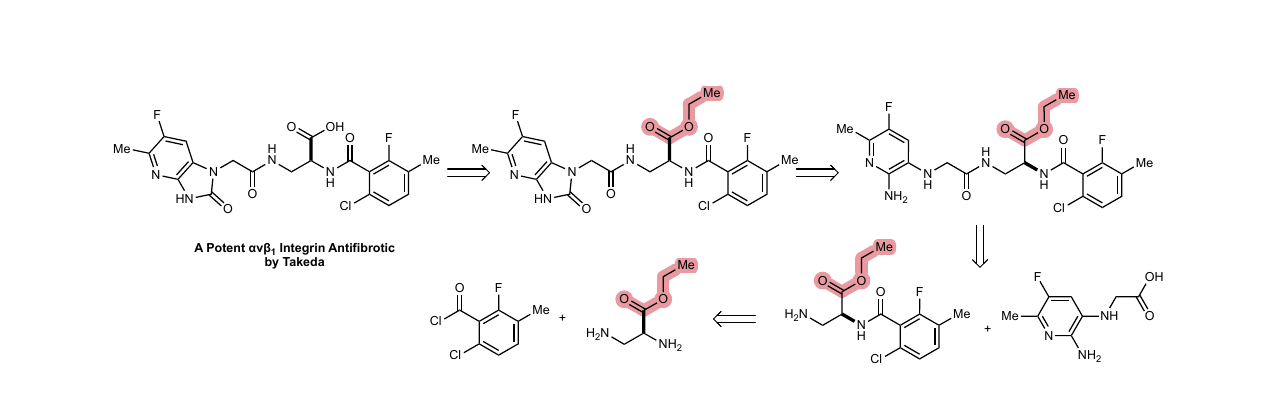}
    \caption{Case study on protecting groups 1.}
    \label{fig:pg_i}
\end{figure}

In this experiment, we examined routes where retrosynthetic planning tools proposed unnecessary protecting groups. The route is shown in \ref{fig:pg_i}. Claude-3.5-Sonnet successfully identified an unnecessarily protected ethyl ester carboxylic acid that was carried through multiple steps and removed in the final step. The model's analysis demonstrated sophisticated chemical reasoning, correctly determining that:
\begin{itemize}
\item The initial amide bond formation could proceed selectively without protection due to the substantially higher nucleophilicity of amines compared to carboxylic acids \citep{gromek2016synthesis}
\item The penultimate phosgene-driven amide bond ring synthesis did not require carboxylic acid protection due to kinetic and entropic advantages inherent in five-membered ring formation \citep{english1945studies, clark1958synthesis}
\end{itemize}

\begin{figure}[ht!]
    \centering
    \includegraphics[width=\linewidth]{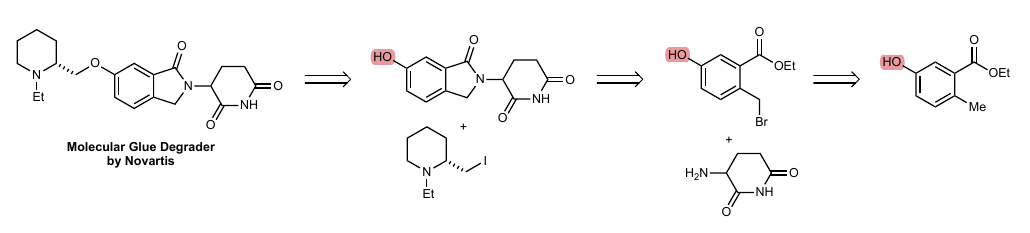}
    \caption{Case study on protecting groups 2.}
    \label{fig:pg_ii}
\end{figure}

In a second experiment, we investigate the inverse, that is flagging routes which require protecting groups at some stage, but don't have them. The model correctly flagged a reactive hydroxyl group that could potentially undergo unwanted intramolecular polymerization with a bromide group elsewhere in the molecule. Furthermore, the LLM proposed appropriate protection strategies, suggesting either TBS (tert-butyldimethylsilyl) or MOM (methoxymethyl) protecting groups. This is demonstrated in \ref{fig:pg_ii}

\clearpage
\section{Strategy-aware benchmark targets}
\label{si:steering-prompts}

The specific prompts and molecular targets used for the strategy-aware synthesis planning benchmark are listed here:

\begin{figure}[ht!]
    \centering
    \includegraphics[width=0.5\linewidth]{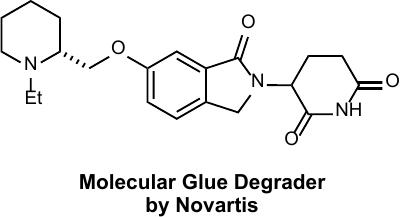}
\end{figure}

\begin{itemize}
    \item[A:] Break piperidine and oxoisoindolinone rings in the synthesis. Get the piperidine-2,6-dione from commercially available materials.
    \item[B:] Break piperidine-2,6-dione and oxoisoindolinone rings in the retrosynthesis. Get the other piperidine ring from commercially available materials.
    \item[C:] Break only oxoisoindolinone ring in synthesis. Get piperidine-2,6-dione and piperidine rings from commercially available materials.
\end{itemize}

\begin{figure}[ht!]
    \centering
    \includegraphics[width=0.5\linewidth]{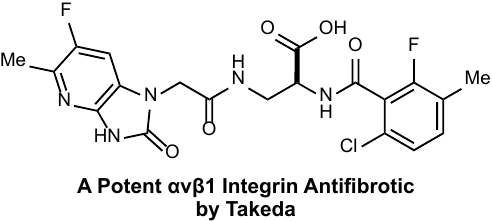}
\end{figure}

\begin{itemize}
    \item[D:] No ring formation reaction.
    \item[E:] Late imidazole ring formation.
    \item[F:] Early imidazole ring formation.
\end{itemize}

\begin{figure}[ht!]
    \centering
    \includegraphics[width=0.5\linewidth]{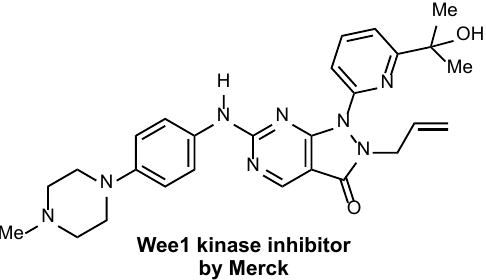}
\end{figure}
\begin{itemize}
    \item[G:] Don't break any ring but get all rings from commercial materials.
    \item[H:] Break pyrimidine in the early stage but get all other rings from commercially available materials.
\end{itemize}

\begin{figure}[ht!]
    \centering
    \includegraphics[width=0.9\linewidth]{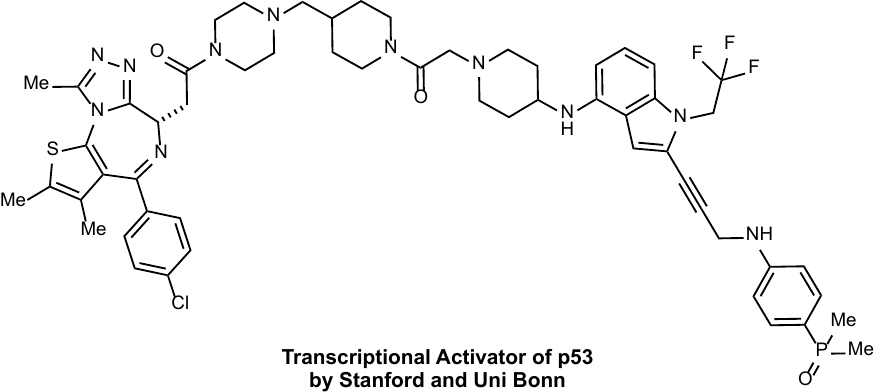}
\end{figure}
\begin{itemize}
    \item[I:] Identify the disconnection strategy that will cut the molecule in two similarly sized intermediates. The disconnection should be made between two piperidine rings.
    \item[J:] Identify the disconnection strategy where the key disconnection will be made between indole and amino-piperidine rings.
    \item[K:] Identify the disconnection strategy that will cut the molecule in two similarly sized intermediates. The disconnection should be made between piperazine and piperidine rings.
    \item[L:] Identify the disconnection strategy that will cut the molecule in two similarly sized intermediates. One intermediate will have piperidine, indole, and aniline rings. The other intermediate will have thiophenol, chlorobenzene, diazepine, triazole, piperazine, and the other piperidine rings.
    \item[M:] Identify the disconnection strategy that will cut the molecule in two intermediates. The disconnection should be made between diazepine and piperazine rings.
\end{itemize}

\clearpage
\section{Analysis of alternatives for Strategy-aware Synthesis Planning}
\label{app:alternative-approaches}

We compare our proposed LLM-guided strategy-aware retrosynthesis framework with potential alternative approaches for tackling the filtering or scoring synthetic routes according to user-specified requirements. The approach we present in this work enables users to express arbitrary preferences in natural language, with LLMs evaluating route alignment in an end-to-end, user-friendly manner. Similar functionality can be achieved using traditional cheminformatics, custom code, and existing or custom ML pipelines. Such alternatives however suffer from adaptability and flexiblity constraints. Here we analyze the practical requirements, applicability and limitations for several alternative strategies in contrast to our method.

We structure this section as follows: We propose 4 case-studies, each represented by an input set (molecular target, prompt). After this, an explanation follows on what it would take to reproduce a similar behavior using other currently existing tools, and we finalize with a table that summarizes the level of effort required in several areas.

\clearpage
\subsection{Case study 1}

\textbf{Prompt:} Late imidazole ring formation.

\begin{figure}[ht!]
    \centering
    \includegraphics[width=0.5\linewidth]{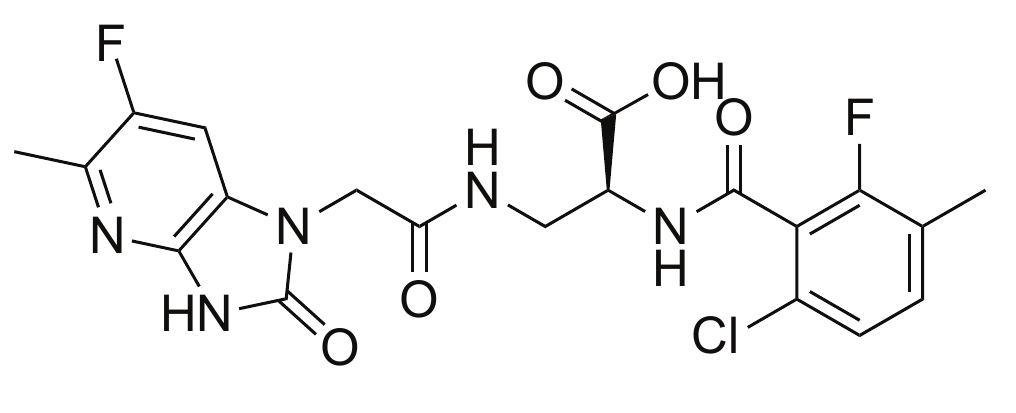}
\end{figure}

Tackling this prompt requires mainly 3 steps: (1) identifying the relevant structural motif (imidazole), (2) identifying whether a reaction exists in the route, where that motif is formed, (3) judging if the relative positioning of such reaction classifies it as "late".

All these steps could be tackled with e.g. Python and RDKit, the full workflow is outlined below:

\begin{itemize}
    \item Substructure SMARTS matching (e.g., \texttt{[nH]1cncc1}) to identify imidazole motifs in each intermediate of each route
    \item For every reaction in the route, check for the appearance of the imidazole ring in the products, not present in the reactants. If so, store the reaction ID and depth of the route where this happens.
    \item Define "late" (e.g. as percentage of route progress)
\end{itemize}

Note that this workflow does not account for edge cases. For instance, if the imidazole ring comes from a different ring (e.g. an imidazoline) through a functional group interconversion, this would be classified as relevant by the alternative approach, which is technically correct but most likely not what is meant by the user when and "imidazole ring formation" is requested.

\begin{table}[ht!]
\begin{tabular}{p{3.6cm}p{4.0cm}p{5.3cm}}
        \toprule
\textbf{Method}   & \textbf{LLM-guided (ours)}  & \textbf{Alternative workflow}      \\
        \midrule
Rule-based cheminformatics & \multirow{3}{*}{Not required}  & Custom SMARTS matching  \\
     &                   & Reaction parsing          \\
     &                   & Route traversal           \\
    \midrule
Custom ML models   &   \multirow{3}{*}{Not required}   & Not required    \\                             &                & Needs dataset of imidazole ring formations\\
                   &                & Retrain/new classifier per motif/query \\
    \midrule
Commercial software        & Not required   & No single tool to handle such ad hoc query flexibly     \\
    \midrule
Robustness/Flexibility  & Handles ambiguities in user intent, flexible   & Brittle. Definition of "formation", "late" must be hardcoded  \\ 
    \bottomrule
\end{tabular}
   \caption{Comparison of route filtering by expert-specified query for "Late imidazole ring formation" using our LLM-based workflow versus standard cheminformatic approaches.}
\end{table}

\clearpage
\subsection{Case study 2.}

\textbf{Prompt:} Break diazepine, thiophene, triazole, and piperazine rings. Get the other rings in the synthesis from commercially available materials. After forming the piperazine ring don't use deprotections, but perform amide coupling with substructure that contains diazepine ring. Form the connection between piperazine and diazepine moieties before proceeding to form fused triazole.

\begin{figure}[ht!]
    \centering
    \includegraphics[width=0.8\linewidth]{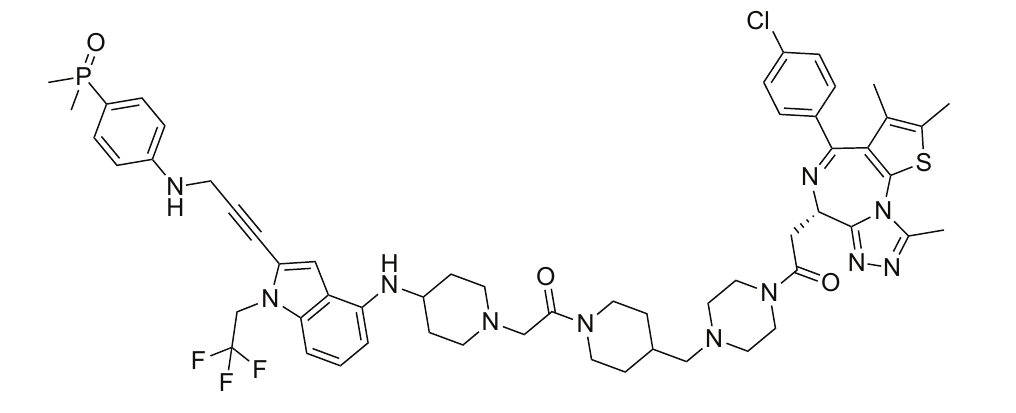}
\end{figure}

Cheminformatics tools would be required to capture the presence and transformation of multiple distinct ring systems (diazepine, thiophene, triazole, piperazine) at each step, requiring SMARTS patterns for identification and routines to detect bond breaking and formation events. The workflow must track the provenance of all other ring systems throughout the route to verify that they are derived from commercially available materials. Ensuring that deprotections are excluded after piperazine ring formation, or that amide coupling with a diazepine precursor follows a given order, would require custom logic for route traversal, dependency tracking, and event sequencing beyond standard toolkits. Managing these requirements for even a single prompt would result in a brittle, highly customized codebase, and each additional synthetic constraint (or change in the query) would necessitate new rule-writing, testing, and validation. Neither existing ML pipelines nor available commercial CASP tools support such high-level, context-dependent multi-objective logic without significant manual curation or scripting, sharply limiting their ability to generalize to new prompts of this complexity.

\begin{table}[ht!]
\begin{tabular}{p{3.6cm}p{4.0cm}p{5.3cm}}
    \toprule
    \textbf{Method}   & \textbf{LLM-guided (ours)}  & \textbf{Alternative workflow}      \\
    \midrule
    Rule-based cheminformatics  & \multirow{4}{*}{Not required}  & Write SMARTS patterns for diazepine, thiophene, triazole, and piperazine rings \\
        &               & Detect formation/breaking of each ring at stepwise level \\
        &               & Identify all other ring systems and check their origin in the route \\
        &               & Implement logic for sequence/order of specific transformations (\textit{after} piperazine, \textit{before} fused triazole, etc) \\
    \midrule
    Custom ML models &  Not required  & Not required \\
    \midrule
    Commercial software        & Not required   & No single tool to handle such ad hoc query flexibly    \\
    \midrule
    Robustness/Flexibility  
        & Captures sequence and compositional logic in natural language; adapts to subtle/overlapping criteria 
        & Each logic constraint must be formally coded, sensitive to definition errors and combinatorial explosion with query complexity \\
    \bottomrule
\end{tabular}
    \caption{Comparison of workflow required to filter for the complex prompt: "Break diazepine, thiophene, triazole, and piperazine rings. Get the other rings in the synthesis from commercially available materials. After forming the piperazine ring don't use deprotections, but perform amide coupling with substructure that contains diazepine ring. Form the connection between piperazine and diazepine moieties before proceeding to form fused triazole."}
\end{table}

\clearpage

\section{Extended benchmarking results}
\label{app:more-performances}

\begin{figure}[ht!]
\label{fig:more-results}
    \centering
    \includegraphics[width=\linewidth]{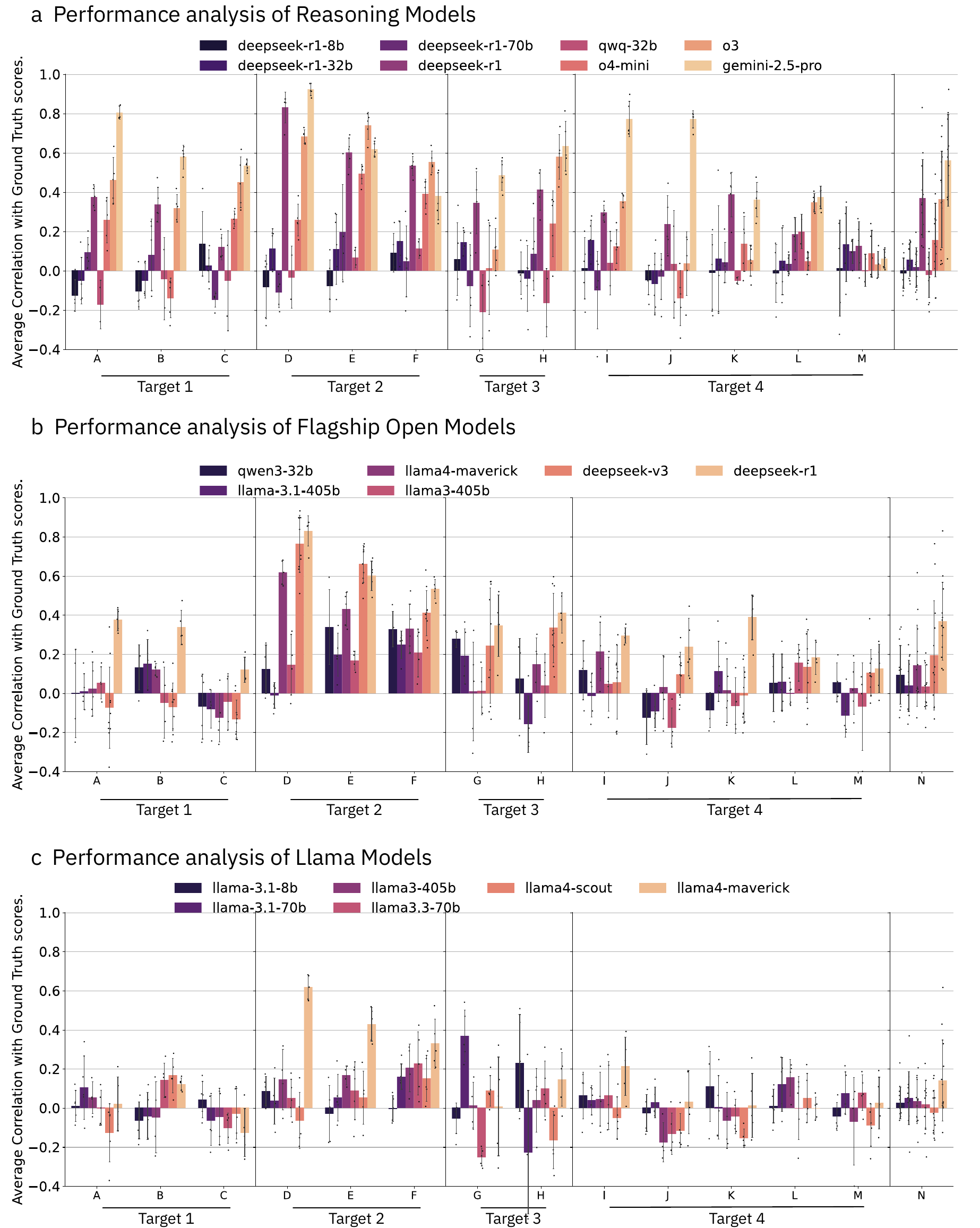}
\end{figure}

We experimented with a wide variety of LLMs to assess the capabilities of different categories of LLMs, including open and closed, as well as small and large models. Additionally, we also tested different models trained specifically for reasoning, such as DeepSeek-r1 models and OpenAI's o3 and o4-mini, with the goal of obtaining a thorough understanding of the achievable capabilities with the current state-of-the-art models. The results are shown in Figure \ref{fig:more-results}.

Figure \ref{fig:more-results}a shows the detailed performances for several reasoning models. As can be seen, gemini-2.5-pro is the leading model, followed by o3 and the open-weights DeepSeek-r1. Smaller versions of the latter do not perform nearly as good, and this might be due to the fact that they are distilled versions of the larger model and have more specific strong capabilities in other areas. The same observation applies to o4-mini, a smaller version of a stronger (to this date unreleased) model, o4. These observations about smaller models further confirm the hypothesis that the chemical analysis capabilities needed for tackling the tasks proposed in this paper are available only beyond a certain threshold in model size. This question remains still open however, and it is left for future work to see if fine-tuned LLMs tailored for this specific task perform as good, or even better than commercial, general LLMs.

The picture for open models, in Figure \ref{fig:more-results}b, looks promising thanks to DeepSeek-r1, however the rest of open models evaluated here did not display strong enough performance to be considered useful for the purposes of our framework. It is still an open question if fine tuning of these open models can unlock new capabilities in open models, that can get close or even surpass the leading LLMs, on the specific tasks relevant to this work.



\clearpage
\section{Evolution of LLM capability over the years}
\label{app:full-evolution}

Reasoning and problem-solving capabilities have been assessed in LLMs since early years of development. Previous research has consistently demonstrated that these capabilities exhibit two key characteristics: first, they tend to emerge at certain model size thresholds rather than appearing gradually \cite{wei2022emergent}, and second, they scale predictably with model size and training compute \cite{brown2020language, kaplan2020scaling}. Studies on mathematical reasoning \cite{lewkowycz2022solving}, code generation \cite{chen2021evaluating}, and multi-step problem solving \cite{suzgun2022challenging} have all shown similar patterns where performance remains relatively flat until a critical model size is reached, after which capabilities improve dramatically.

\begin{figure}[ht!]
    \centering
    \includegraphics[width=\linewidth]{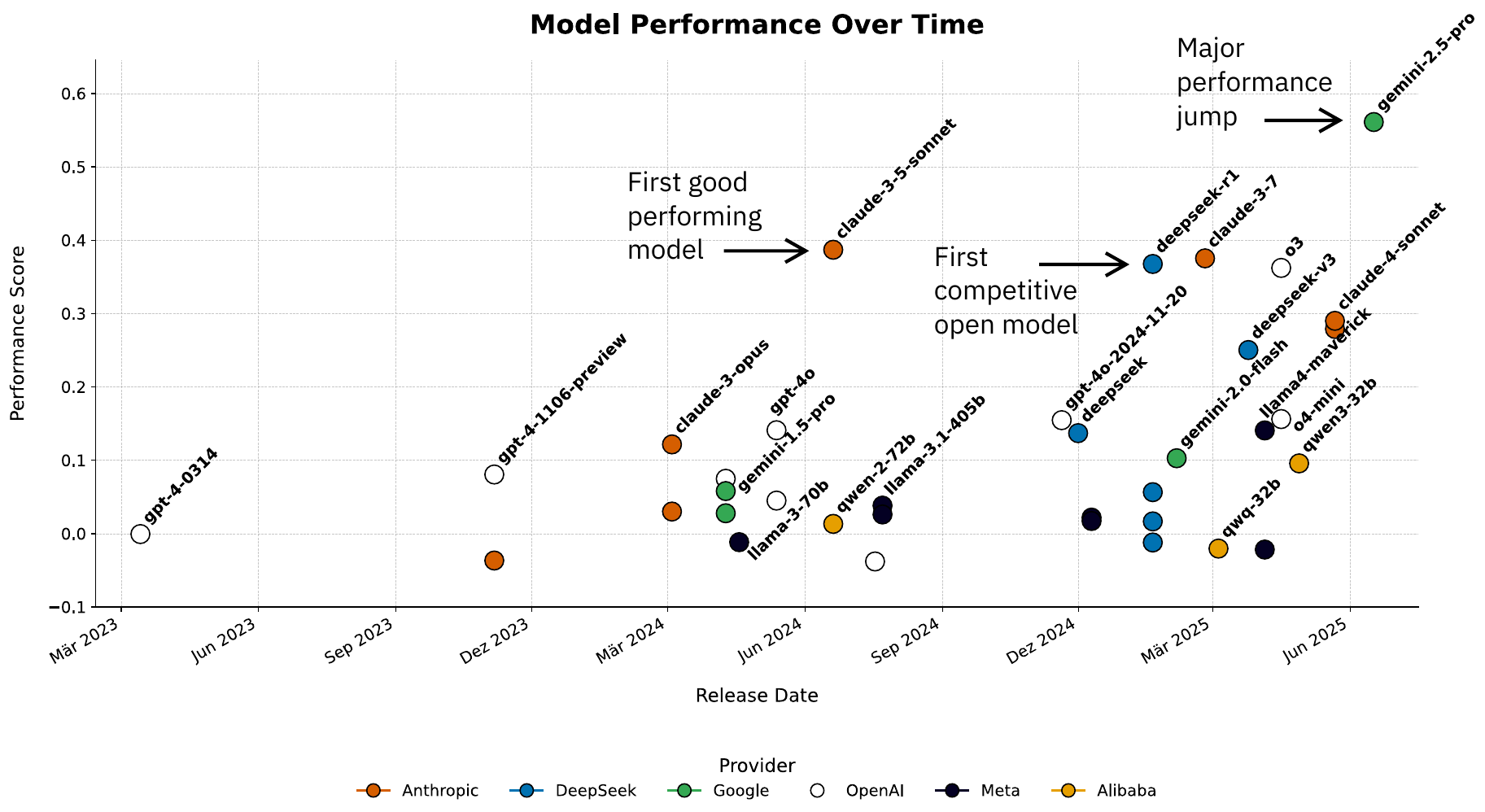}
    \label{fig:full-evolution}
\end{figure}

Our results on chemical analysis tasks follow these established trends, as illustrated in Figure \ref{fig:full-evolution}. The temporal progression of model performance shows a clear upward trajectory, with more recent models consistently outperforming their predecessors. Notably, even the weakest models released in 2025 achieve performance levels that surpass the state-of-the-art models from 2023, demonstrating the rapid pace of improvement in this field. This improvement is particularly pronounced for models released after 2024, coinciding with the introduction of more reasoning-focused training techniques.

The figure also reveals that the performance gains are not merely incremental but often substantial, with new model families showing significant jumps in capability. This pattern suggests that the underlying improvements in model architecture, training methodologies, and scale continue to unlock new levels of reasoning ability, even for specialized tasks like chemical analysis that were not explicitly targeted during training. The consistent improvement across different model families and organizations indicates that these advances represent genuine progress in fundamental reasoning capabilities rather than task-specific optimizations.

\clearpage
\section{Sample model responses}
\label{app:some-gemini-outputs}

\begin{figure}[ht!]
    \centering
    \includegraphics[width=\linewidth]{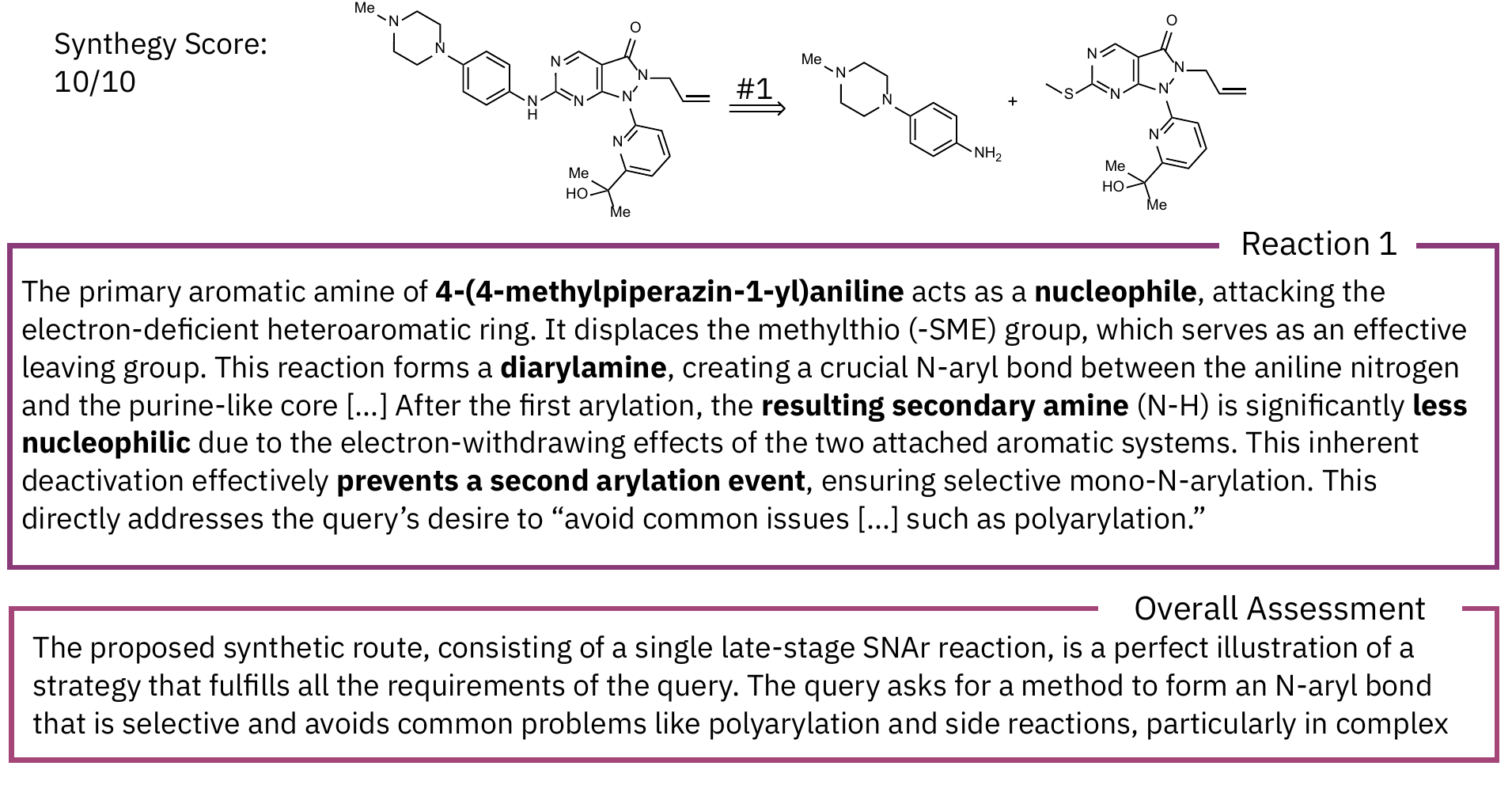}
\end{figure}

\begin{figure}[ht!]
    \centering
    \includegraphics[width=\linewidth]{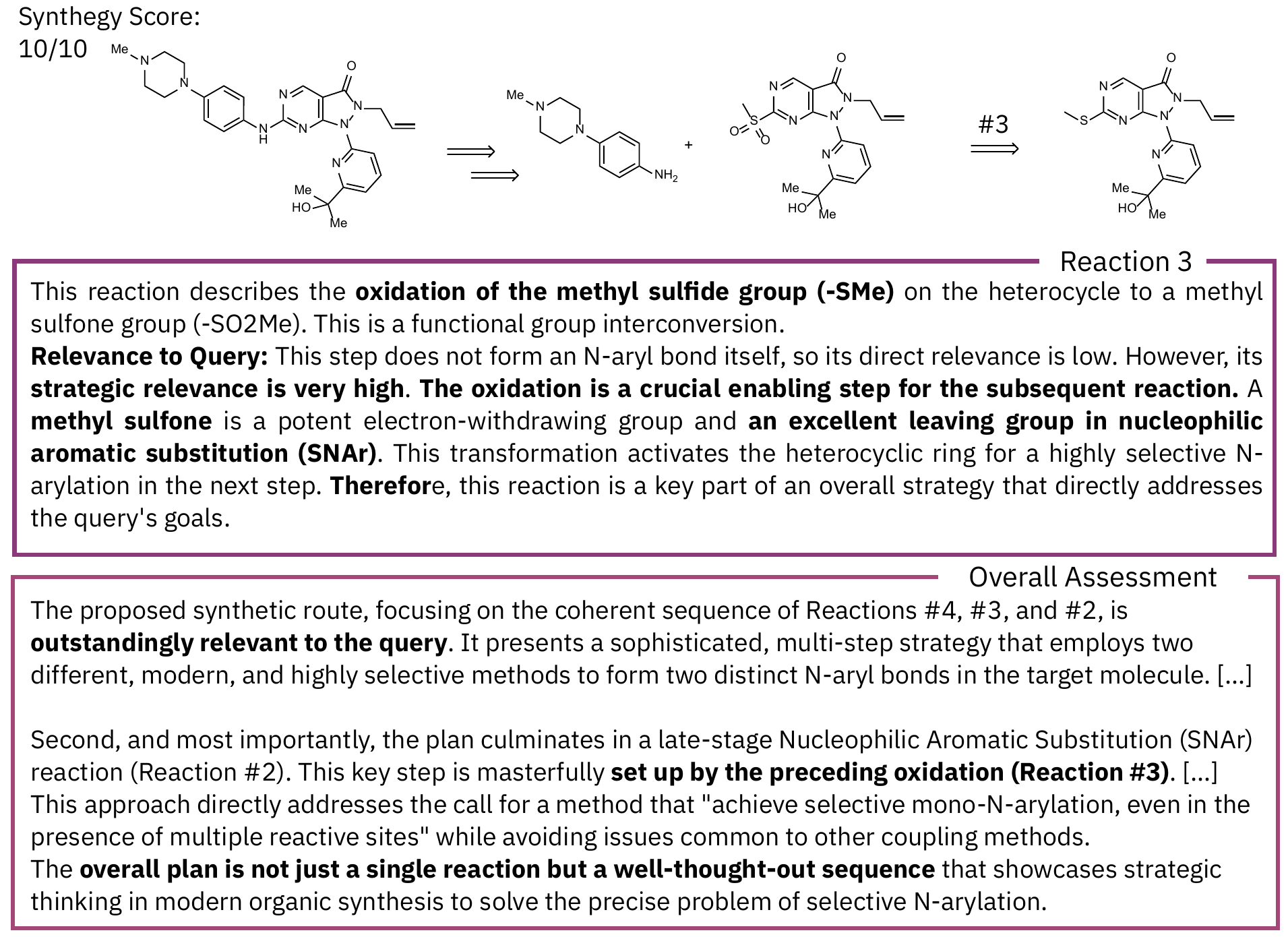}
\end{figure}

\begin{figure}[ht!]
    \centering
    \includegraphics[width=\linewidth]{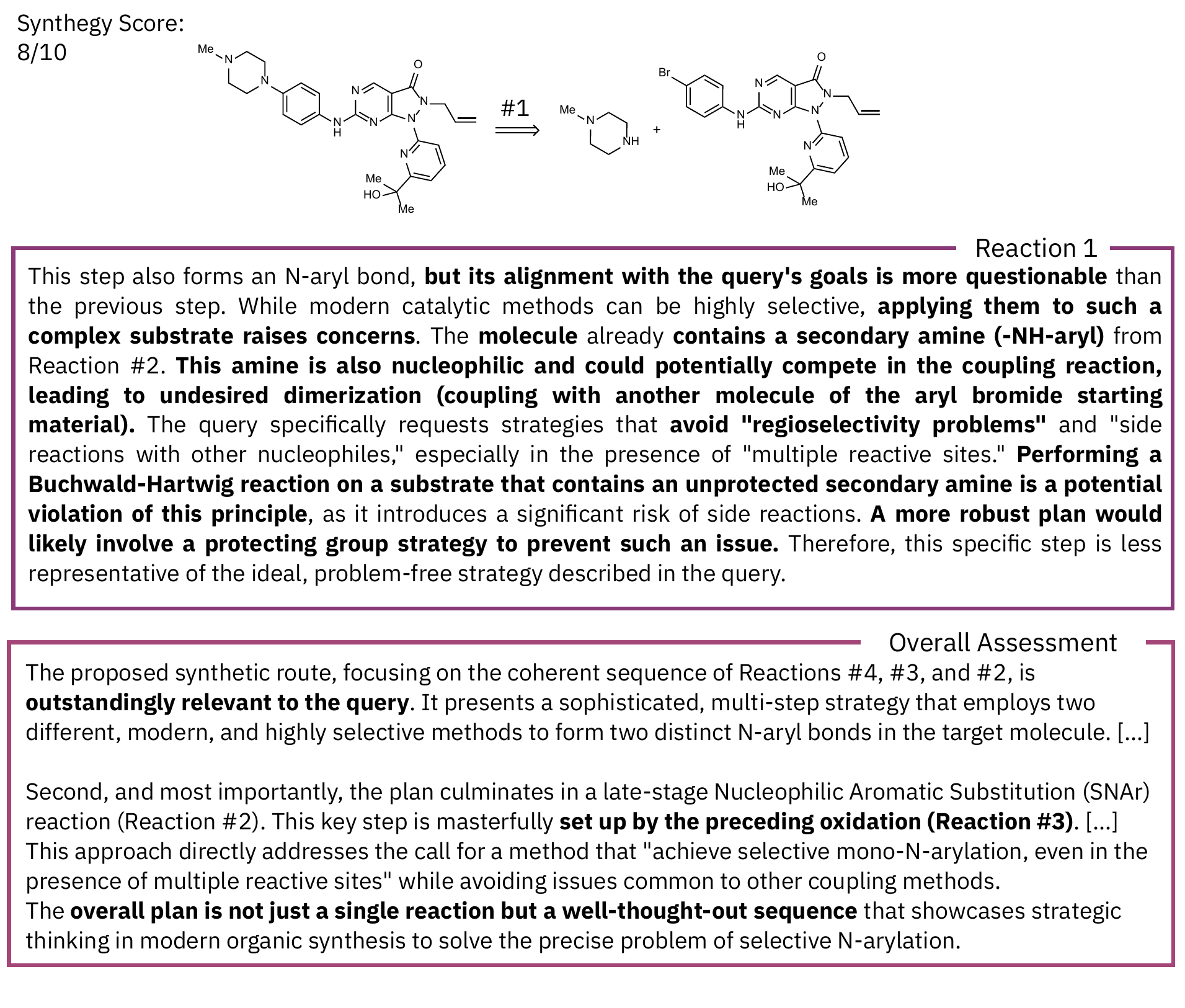}
\end{figure}

\clearpage
\section{Case Studies - Selection of feasible synthesis routes}
\label{app:why-low-score}

In this section we show some examples of Synthegy's capabilities at detecting errors and potential problems in synthetic routes, highlighting single-reaction failures as well as overall inefficiencies among others, leading it to give low scores for those routes.

\begin{figure}[ht!]
    \centering
    \includegraphics[width=\linewidth]{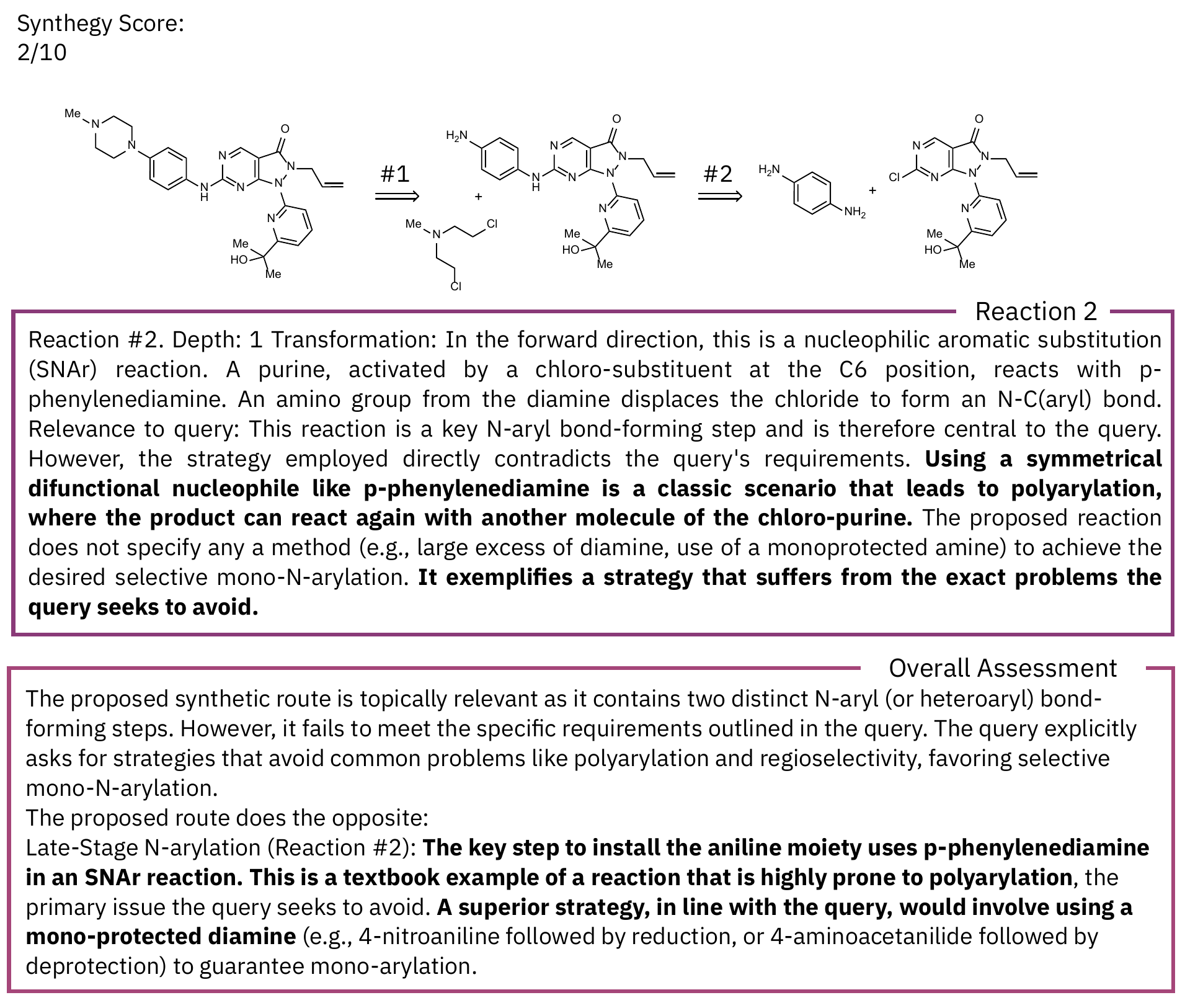}
\end{figure}

\begin{figure}[ht!]
    \centering
    \includegraphics[width=\linewidth]{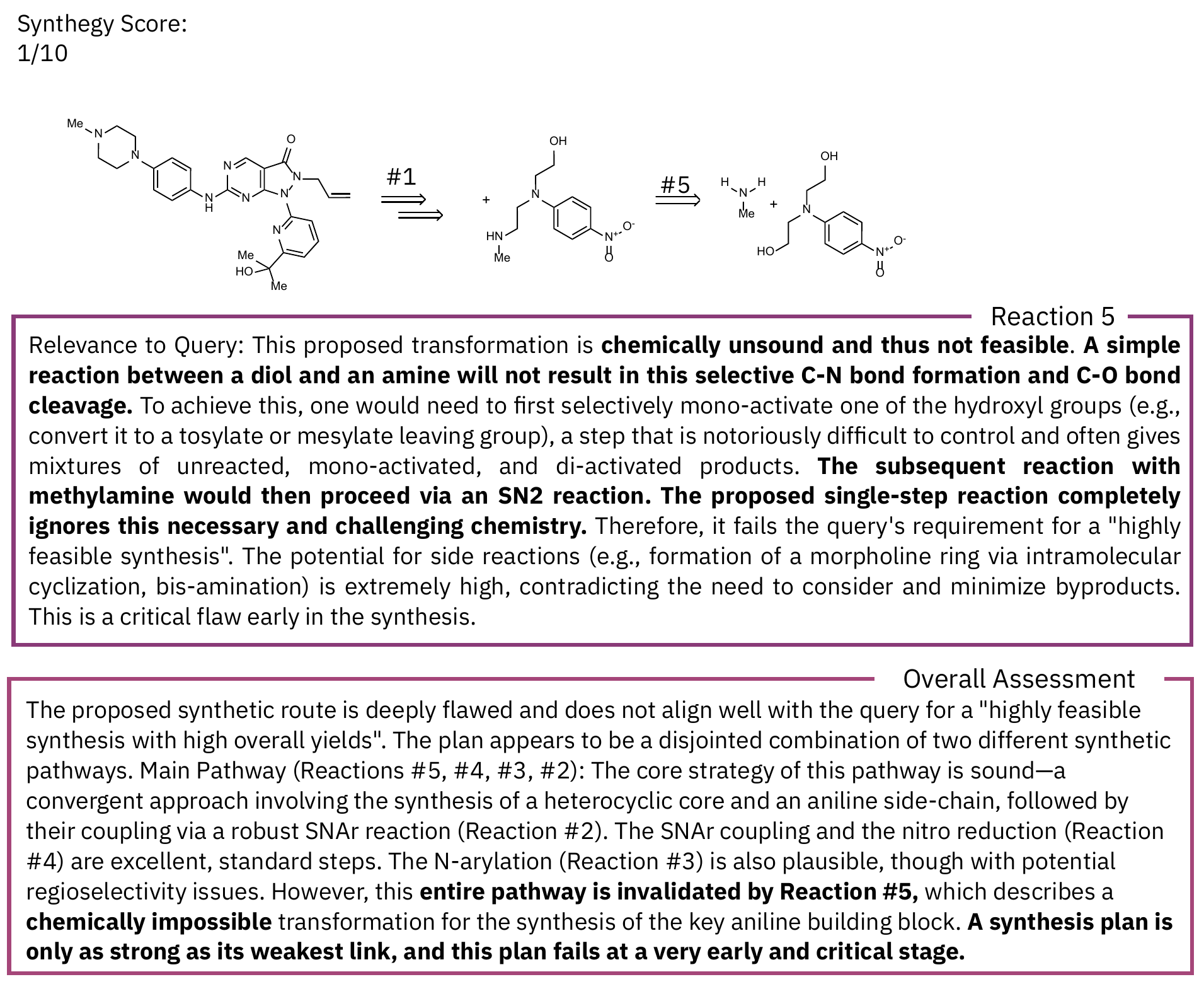}
\end{figure}

\clearpage
\section{Case studies in complex synthetic routes}

To study with more detail the analytical patterns of LLMs when considering synthetic routes, we setup a case study using two molecules for which several routes have been published throughout the years: atorvastatin \cite{atorvastatin_roth1990, atorvastation_process_1993} and strychnine \cite{cannon_is_2012}. We select Claude-3.7-Sonnet for this analysis. We design a set of expert prompts that aim to describe specific strategic details of selected routes \cite{genheden2021clustering}, see SI-\ref{si:prompts-case-study}. Our framework is then used to score each of the routes based on their alignment to each of the prompts, and we evaluate whether LLMs can effectively separate the described route from the rest, given a suitable prompt.

\begin{figure}[ht!]
    \centering
    \includegraphics[width=\linewidth]{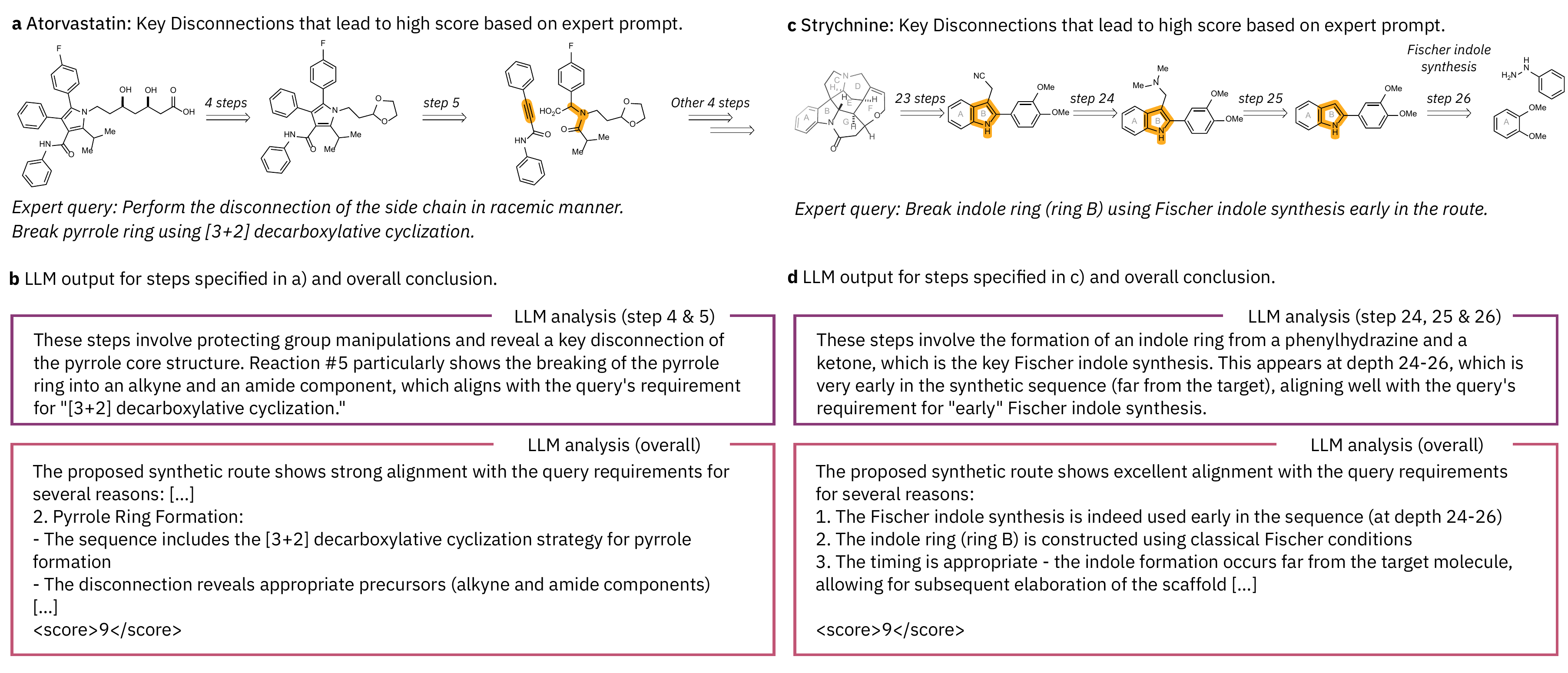}
    \caption{\textbf{High-ranked route examples for complex syntheses.} a) Key Disconnections of atorvastatin that leads to high score based on expert prompt as detected by LLM. b) LLM's reasoning about these transformations, and final analysis to check alignment with expert query and score the route. c) Key Disconnections of strychnine that leads to high score based on expert prompt as detected by LLM. d) Corresponding LLM reasoning.}
    \label{fig:figure_atorvastatin}
\end{figure}

The results for Atorvastatin in Figure \ref{fig:figure_atorvastatin}a show that the LLM is indeed capable of detecting the exact transformation, within the complete retrosynthetic tree, that satisfies the expert query, and consequently scoring this route highly (9/10). As anticipated in the previous section, the LLM follows a strategy of analyzing each reaction individually and in relation with the query, then revising the produced analyses to synthesize a final summary that evaluates the alignment with the query, along with the final score, which we use for evaluation. Among the key features of correct analyses, we find typically correct description of reactions in terms of the names, reaction mechanisms, and overall features of the transformations at hand. When analyzing the key transformation (step 5 in this case), the LLM correctly characterizes the ring formation reaction --- correct mechanistic considerations, emerging functional groups --- and correctly highlights its alignment with the query's requirement (Figure \ref{fig:figure_atorvastatin}b). In this example, it's worth noting that the LLM is successful at navigating a chemical space of more than 30 ring systems across over 10 complex reactions.

We conduct a final stress-test on a molecule whose synthesis has been regarded as one of the key achievements in early organic synthesis, and for which novel synthetic methods are still being developed due to its complexity and academic interest --- strychnine. A higher number, and more complex routes have been published for this archetypal natural product, making it ideal for a comparison of this kind. Our results show that the best LLMs are indeed able to correctly detect different synthetic strategies for several lengthy routes, of over 20 reaction steps. In particular, the example in Figure \ref{fig:figure_atorvastatin}c demonstrates how the system is able to detect and highly score Woodward's strychnine synthesis, from a descriptive strategy-oriented query. The high score in detecting Fischer indole synthesis (Figure  \ref{fig:figure_atorvastatin}d) demonstrates that LLMs can effectively analyze SMILES strings of extended synthetic trees of complex natural products, such as those with 26 reaction steps, while remaining attentive to the original prompt.

These results represent strong evidence that current LLMs are capable of analyzing and interpreting molecules, reactions, and synthetic routes across multiple dimensions, from structural analysis to synthetic strategy. Furthermore, our results show that these capabilities can directly be exploited in combination with traditional search algorithms, further expanding the possibilities of these high-performing systems and, as shown, enabling strategy-aware synthesis planning.

\clearpage
\subsection{Strategic queries used in the case studies}
\label{si:prompts-case-study}

The objective with this case-study is to determine whether, given some specification of a route in terms of desired strategic elements, our framework can selectively yield the correct route, from a set of real routes with historical relevance. Here we present the

\begin{figure}[ht!]
    \centering
    \includegraphics[width=0.7\linewidth]{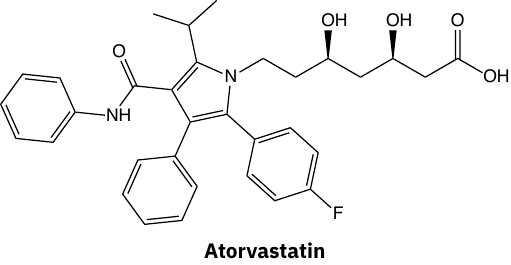}
\end{figure}

\paragraph{Prompts for Atorvastatin:} We use the dataset of historical routes from \cite{genheden2025simple}, contains 3 routes, which can be roughly categorized in terms of the strategic disconnection of the central polysubstituted pyrrole ring: decarboxylative, formal [3+2] cyclization, or through a Paal-Knorr pyrrole synthesis reaction. Prompts were designed to represent these categories as follows.

\begin{enumerate}
    \item Break pyrrole ring relatively early in the retrosynthesis through Paal-Knorr condensation to provide convergent synthesis.
    \item Perform the disconnection of the side chain in racemic manner. Break pyrrole ring using [3+2] decarboxylative cyclization.
    \item Break pyrrole ring using [3+2] decarboxylative cyclization.
\end{enumerate}

\begin{figure}[ht!]
    \centering
    \includegraphics[width=0.4\linewidth]{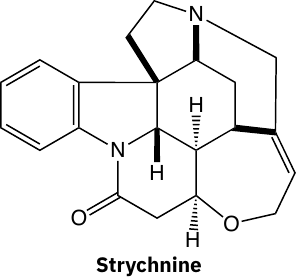}
\end{figure}
\paragraph{Prompts for Strychnine:} We obtain 10 routes from \cite{genheden2025simple} which are adapted from works in the literature. 
The research articles where these routes are presented tend to describe the strategy in better detail, hence our prompts are derived from such publications:

\begin{enumerate}
    \item Break indole ring (ring B) using Fischer indole synthesis early in the route.
    \item Strategically use veratryl group as a source of muconic ester via oxidative cleavage, prior to synthesizing pyridone ring.
\end{enumerate}

\clearpage
\section{Cost and Latency analysis}
\label{app:cost-latency}

A key concern for the deployment of Synthegy in practical synthetic planning workflows is the potential costs and latency associated with LLM usage. To better understand this, we analyze the costs and latency associated with 3 LLMs used in this work: Gemini-2.5-pro, DeepSeek-R1, and Claude-3.5-Sonnet. Results are displayed in Table \ref{tab:costs}. Our analysis demonstrates that the computational requirements for reranking synthetic routes are modest, with individual evaluations completing in seconds to minutes at costs not exceeding 15 USD for the best-performing models. For a typical synthesis requiring the evaluation of 60 candidate routes, the entire reranking process can be completed within 12 minutes at a cost of approximately \$2-3, making it highly accessible for routine use in both academic and industrial settings. Furthermore, costs are dropping rapidly while model performance continues to improve dramatically \cite{aiindexreport}. This trajectory suggests that future applications could be both exceptionally cheap and highly performant, positioning our current work as a very early demonstration of capabilities that will become increasingly powerful and accessible over time.

\begin{table}[ht!]
\centering
\caption{LLM Cost Analysis for strategy-aware synthesis planning. The analysis demonstrates the feasibility of deploying LLM-based reranking for synthetic route optimization. For a single synthesis requiring reranking of 60 candidate routes, the process can be completed in under 12 minutes using Claude-3.5-Sonnet with excellent results at a cost of only \$2.19 per batch. Even the most accurate approach using Gemini-2.5-Pro would cost under \$3.24 per molecular target while maintaining high-quality chemical reasoning, making this technology highly accessible for routine synthetic planning workflows.}
\begin{tabular}{lccc}
\toprule
Model           & \textbf{Gemini-2.5-Pro} & \textbf{DeepSeek-R1} & \textbf{Claude-3.5-Sonnet}\\ 
\midrule
Avg Tokens      & 2,533 / 5,067                       & 4,697 / 9,393                    & 917 / 1,833                            \\ \hline
Cost per Call (\$)   & 0.0540                              & 0.0225                           & 0.0365                                 \\ \hline
Latency per Call (s) & 74                                  & 198                              & 12                                     \\ \hline
Cost per Batch (\$) & 3.24                                & 1.35                             & 2.19                                   \\ \hline
Benchmark Time & 22m                                 & 1h30                             & 5m                                    \\ \hline
Benchmark Cost (\$)  & 12.96                               & 5.40                             & 8.75                                   \\ \hline
Performance (corr)    & 0.56                                & 0.36                             & 0.38                                   \\ \hline
\end{tabular}
\label{tab:costs}
\end{table}

\clearpage
\section{Mechanistic benchmark design}
\label{si:mechbench-rxns}

Here we illustrate the process of predicting elementary moves in a chemical transformation using the LLM as described in the main article. For each reaction, we formulated the associated mechanism using the elementary moves described in the manuscript, this is the ground truth pathway. At each step, we add 5 additional options that represent alternative moves that possibly not lead to the correct product and would typically count as incorrect moves.

\begin{figure}[ht!]
    \centering
    \includegraphics[width=\linewidth]{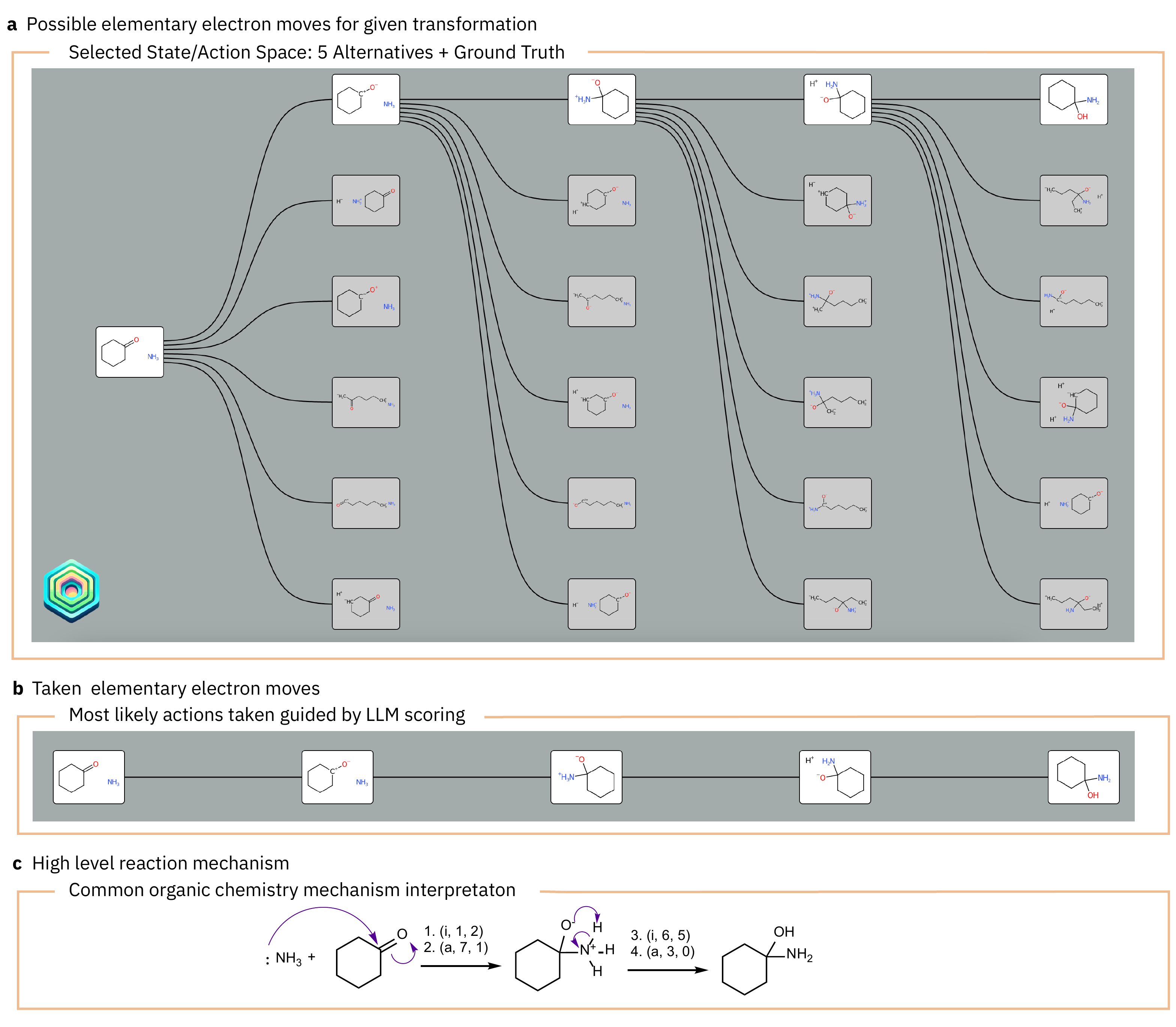}
    \caption{Description of Task 1 in the mechanistic benchmark. \textbf{a} shows the ground truth path (highlighted in white) along 5 other options for each step (light grey). Dark grey indicates moves that are part of the ground truth sequence but have already been traversed, serving as a check against loops in the prediction model. \textbf{b} Isolates the ground truth and \textbf{c} provides an interpretation of the reaction mechanism derived from the ground truth sequence using the electron movement representation.}
\end{figure}
\begin{figure}[ht!]
    \centering
    \includegraphics[width=\linewidth]{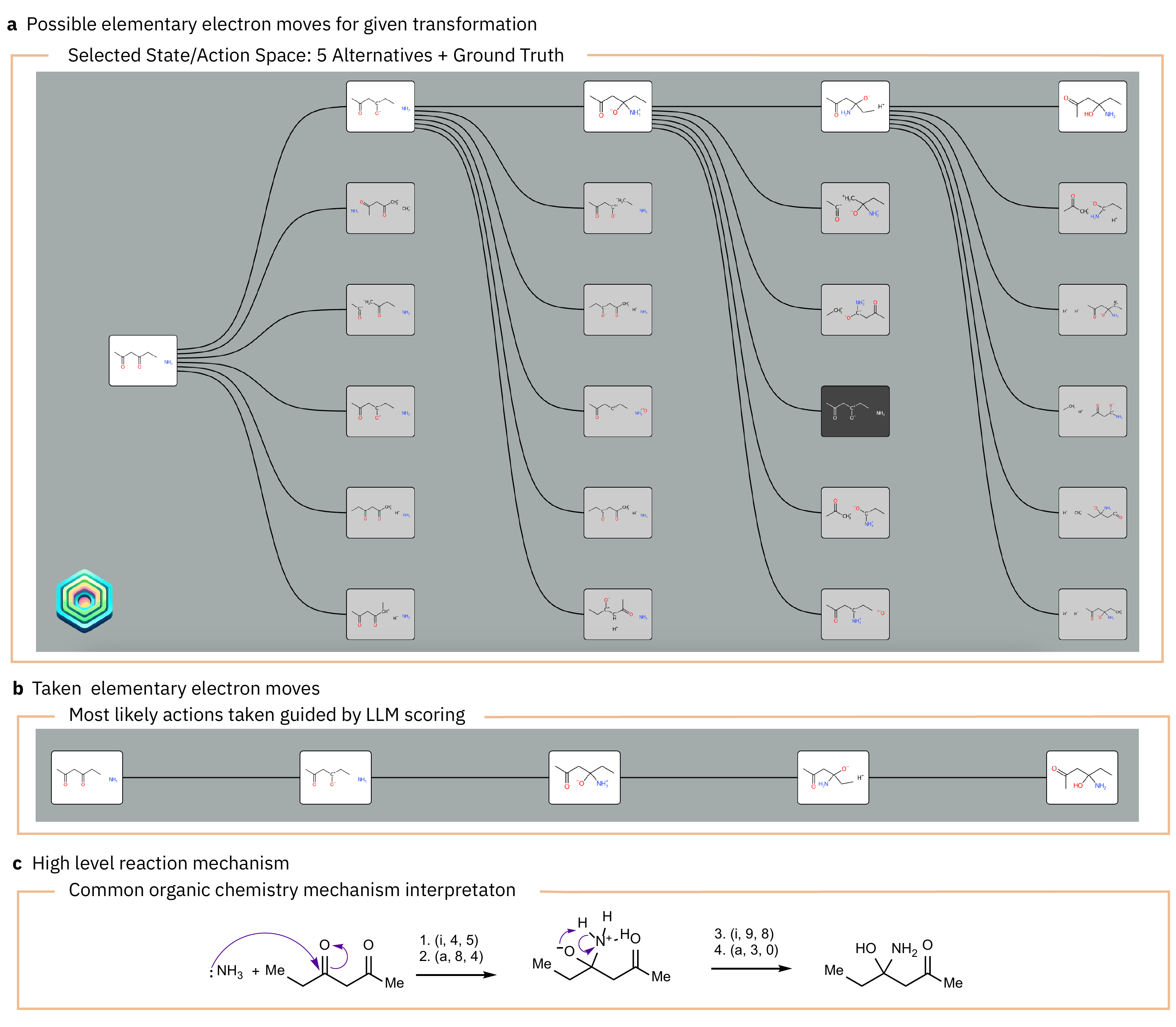}
    \caption{Description of Task 2 in the mechanistic benchmark. \textbf{a} shows the ground truth path (highlighted in white) along 5 other options for each step (light grey). Dark grey indicates moves that are part of the ground truth sequence but have already been traversed, serving as a check against loops in the prediction model. \textbf{b} Isolates the ground truth and \textbf{c} provides an interpretation of the reaction mechanism derived from the ground truth sequence using the electron movement representation.}
\end{figure}
\begin{figure}[ht!]
    \centering
    \includegraphics[width=\linewidth]{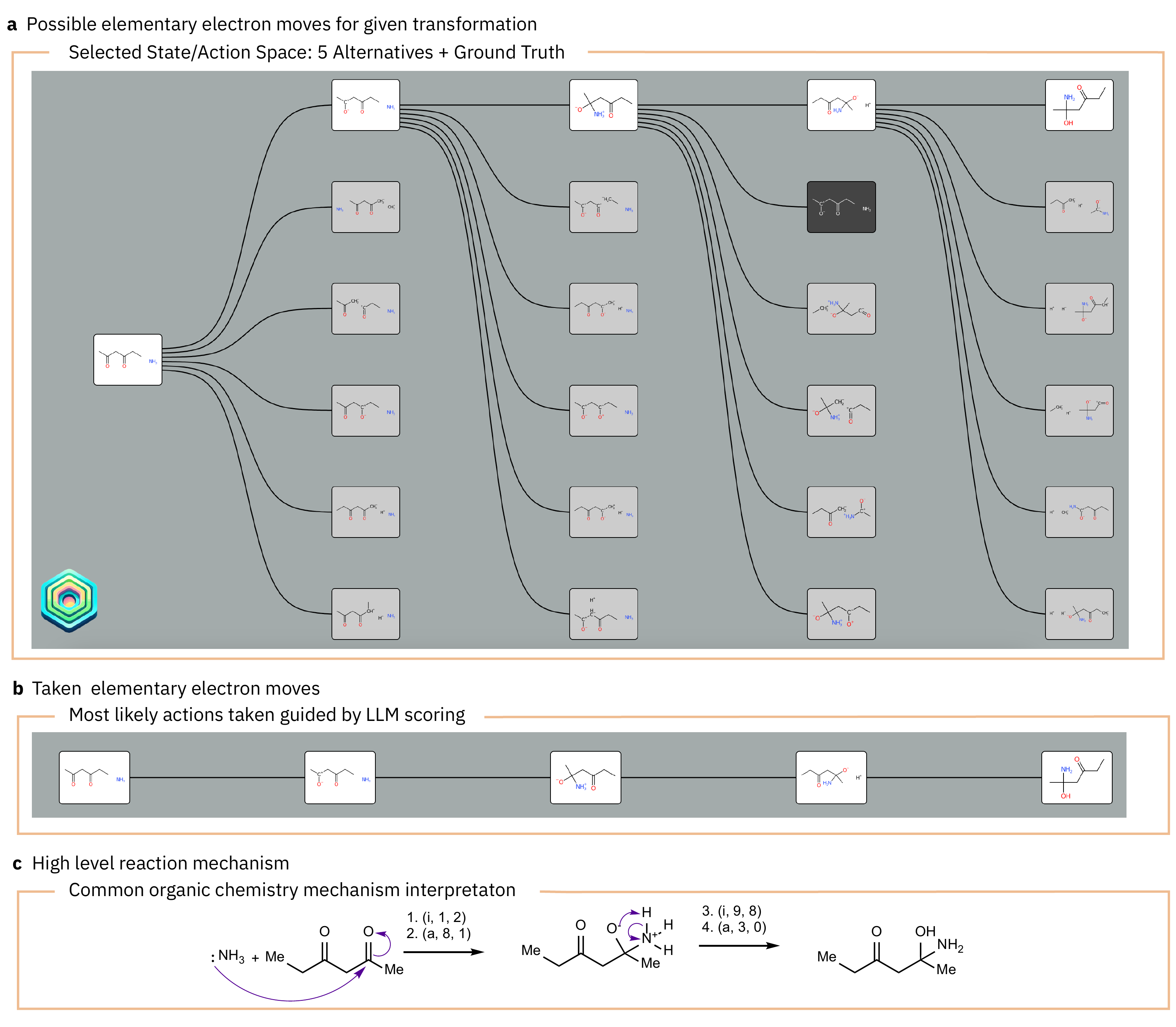}
    \caption{Description of Task 3 in the mechanistic benchmark. \textbf{a} shows the ground truth path (highlighted in white) along 5 other options for each step (light grey). Dark grey indicates moves that are part of the ground truth sequence but have already been traversed, serving as a check against loops in the prediction model. \textbf{b} Isolates the ground truth and \textbf{c} provides an interpretation of the reaction mechanism derived from the ground truth sequence using the electron movement representation.}
\end{figure}
\begin{figure}[ht!]
    \centering
    \includegraphics[width=\linewidth]{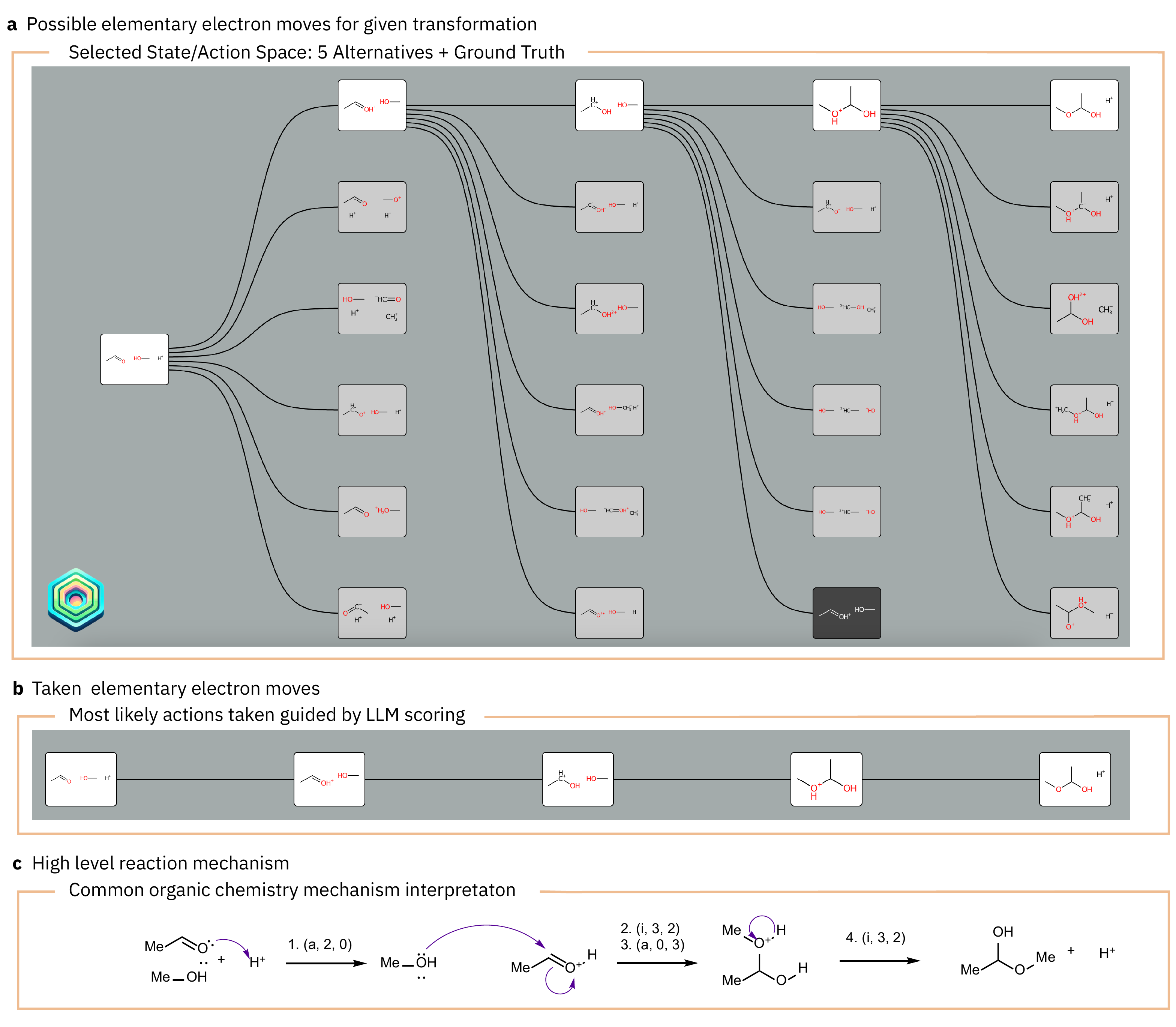}
    \caption{Description of Task 4 in the mechanistic benchmark. \textbf{a} shows the ground truth path (highlighted in white) along 5 other options for each step (light grey). Dark grey indicates moves that are part of the ground truth sequence but have already been traversed, serving as a check against loops in the prediction model. \textbf{b} Shows the isolated ground truth mechanism of the reaction.}
\end{figure}
\begin{figure}[ht!]
    \centering
    \includegraphics[width=\linewidth]{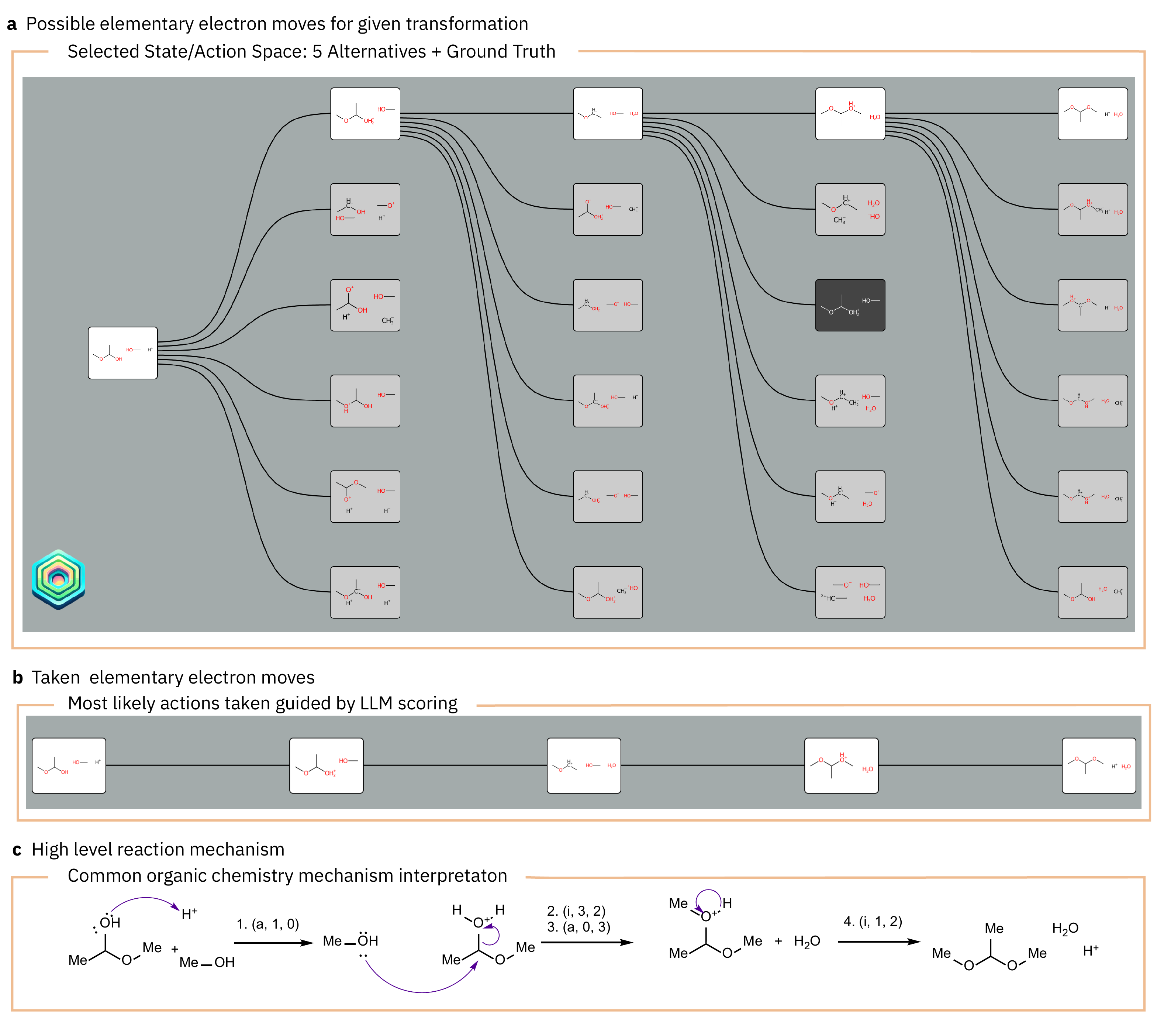}
    \caption{Description of Task 5 in the mechanistic benchmark. \textbf{a} shows the ground truth path (highlighted in white) along 5 other options for each step (light grey). Dark grey indicates moves that are part of the ground truth sequence but have already been traversed, serving as a check against loops in the prediction model. \textbf{b} Shows the isolated ground truth mechanism of the reaction.}
\end{figure}
\begin{figure}[ht!]
    \centering
    \includegraphics[width=\linewidth]{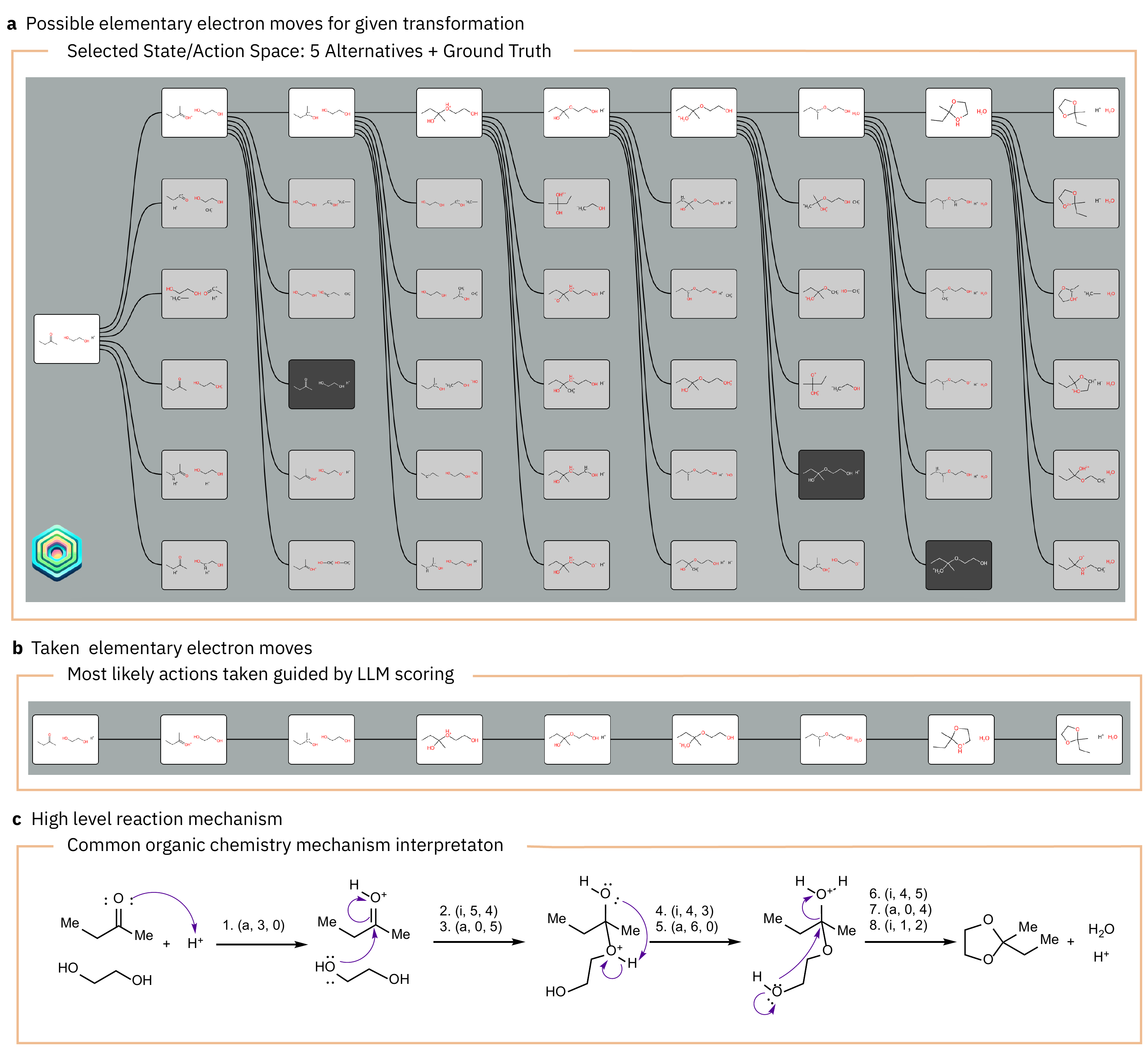}
    \caption{Description of Task 6 in the mechanistic benchmark. \textbf{a} shows the ground truth path (highlighted in white) along 5 other options for each step (light grey). Dark grey indicates moves that are part of the ground truth sequence but have already been traversed, serving as a check against loops in the prediction model. \textbf{b} Shows the isolated ground truth mechanism of the reaction.}
\end{figure}
\begin{figure}[ht!]
    \centering
    \includegraphics[width=\linewidth]{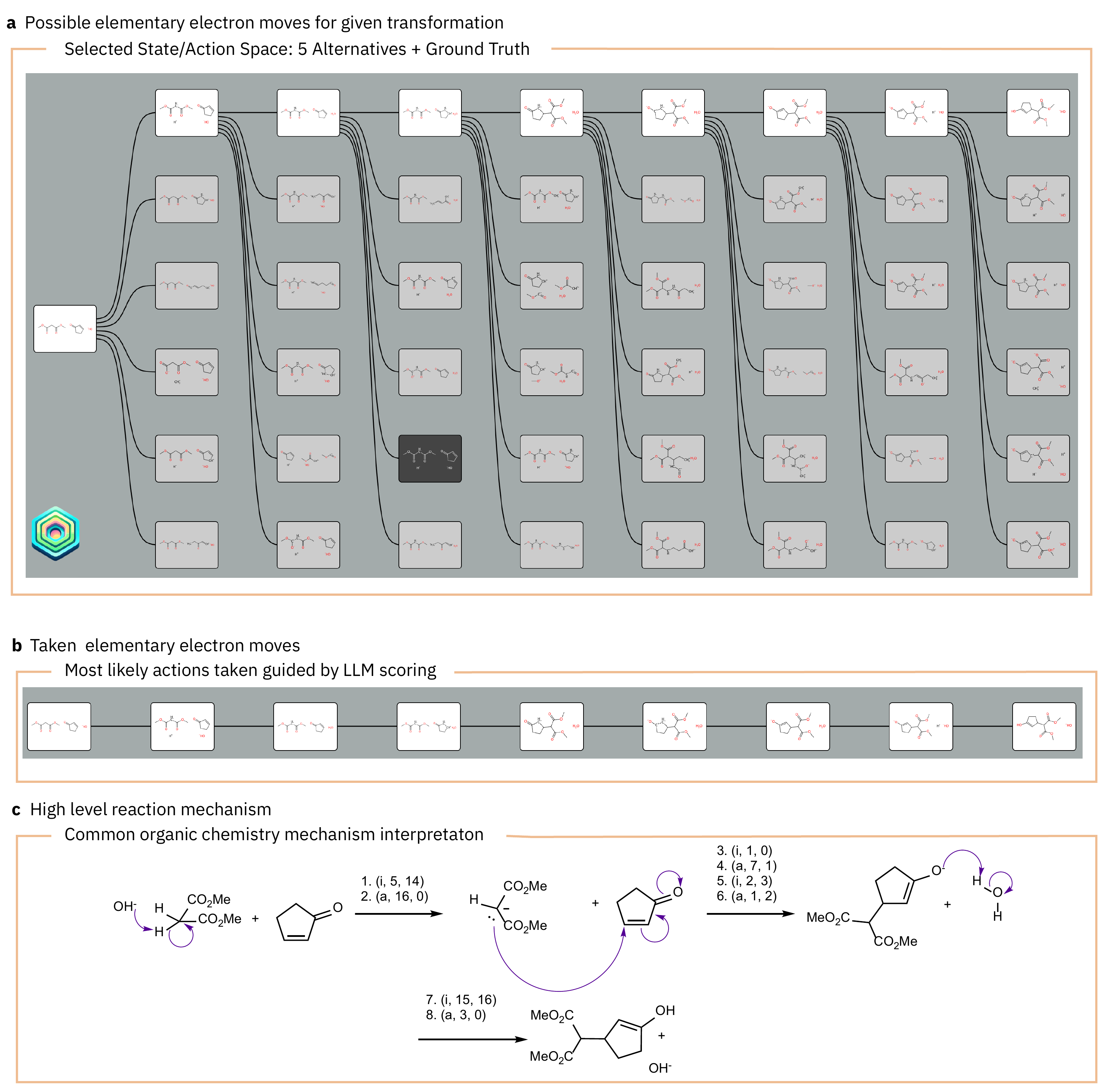}
    \caption{Description of Task 7 in the mechanistic benchmark. \textbf{a} shows the ground truth path (highlighted in white) along 5 other options for each step (light grey). Dark grey indicates moves that are part of the ground truth sequence but have already been traversed, serving as a check against loops in the prediction model. \textbf{b} Shows the isolated ground truth mechanism of the reaction.}
\end{figure}
\begin{figure}[ht!]
    \centering
    \includegraphics[width=\linewidth]{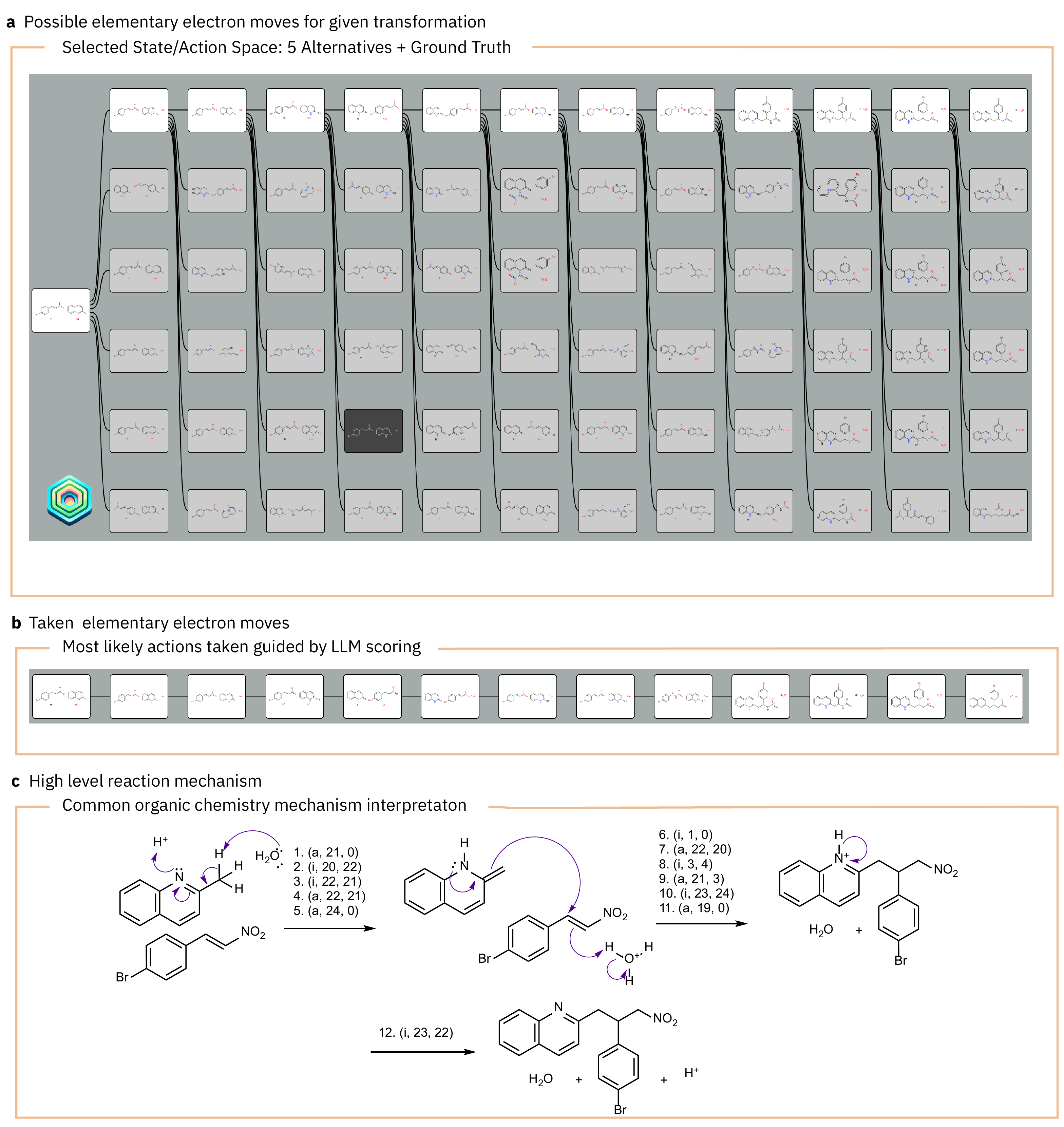}
    \caption{Description of Task 8 in the mechanistic benchmark. \textbf{a} shows the ground truth path (highlighted in white) along 5 other options for each step (light grey). Dark grey indicates moves that are part of the ground truth sequence but have already been traversed, serving as a check against loops in the prediction model. \textbf{b} Shows the isolated ground truth mechanism of the reaction.}
\end{figure}
\begin{figure}[ht!]
    \centering
    \includegraphics[width=\linewidth]{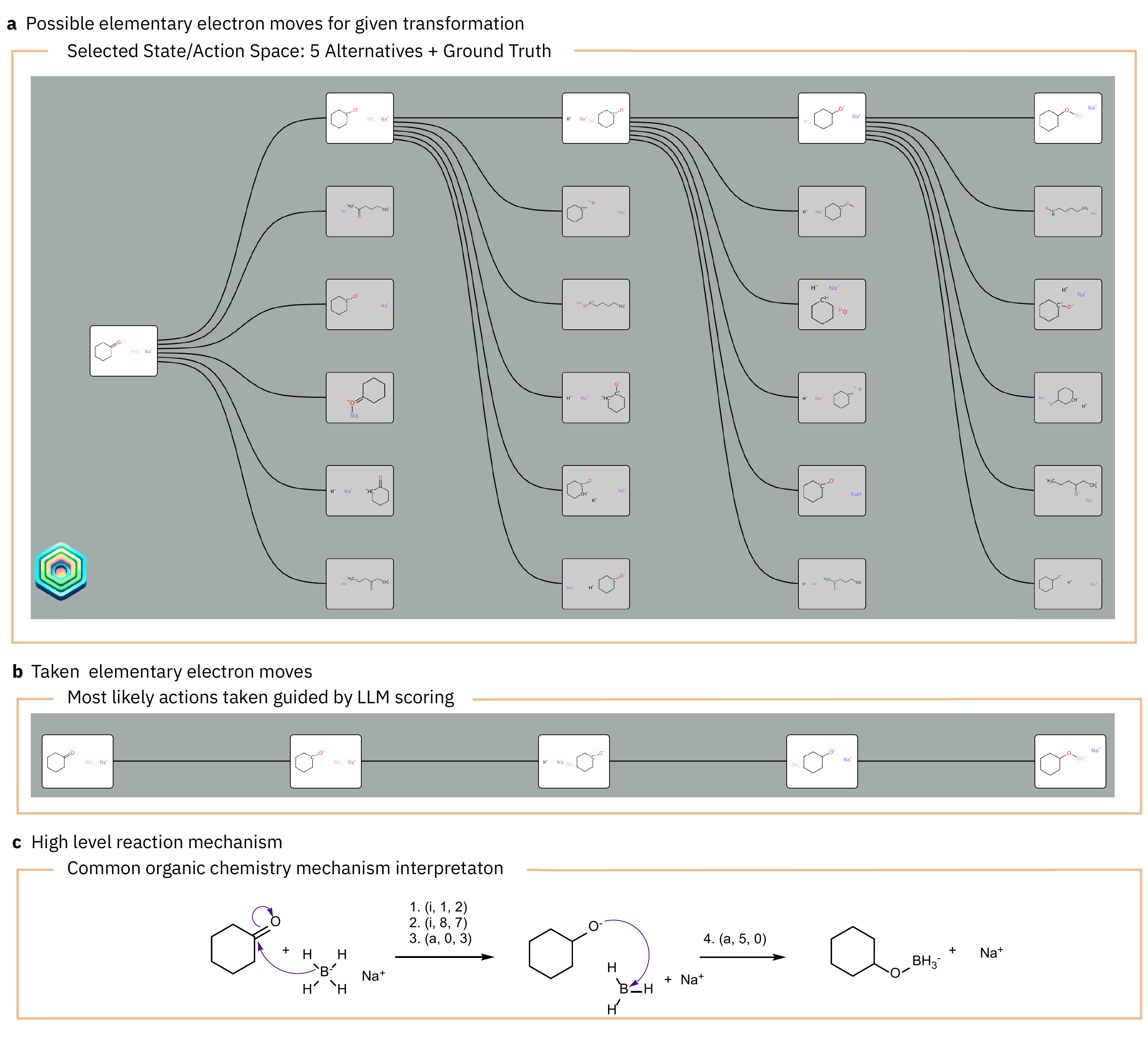}
    \caption{Description of Task 9 in the mechanistic benchmark. \textbf{a} shows the ground truth path (highlighted in white) along 5 other options for each step (light grey). Dark grey indicates moves that are part of the ground truth sequence but have already been traversed, serving as a check against loops in the prediction model. \textbf{b} Shows the isolated ground truth mechanism of the reaction.}
\end{figure}
\begin{figure}[ht!]
    \centering
    \includegraphics[width=\linewidth]{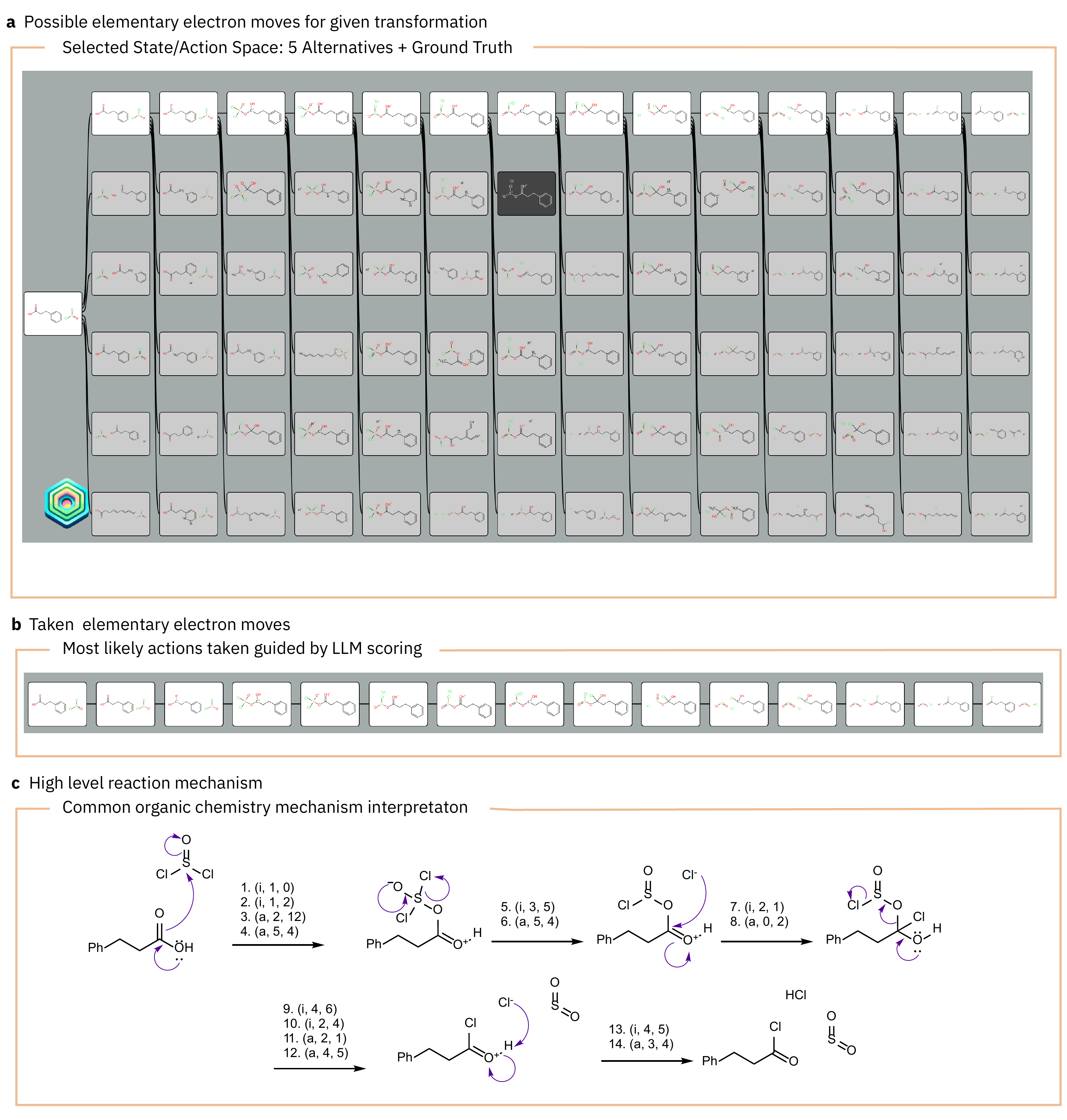}
    \caption{Description of Task 10 in the mechanistic benchmark. \textbf{a} shows the ground truth path (highlighted in white) along 5 other options for each step (light grey). Dark grey indicates moves that are part of the ground truth sequence but have already been traversed, serving as a check against loops in the prediction model. \textbf{b} Shows the isolated ground truth mechanism of the reaction.}
\end{figure}
\begin{figure}[ht!]
    \centering
    \includegraphics[width=\linewidth]{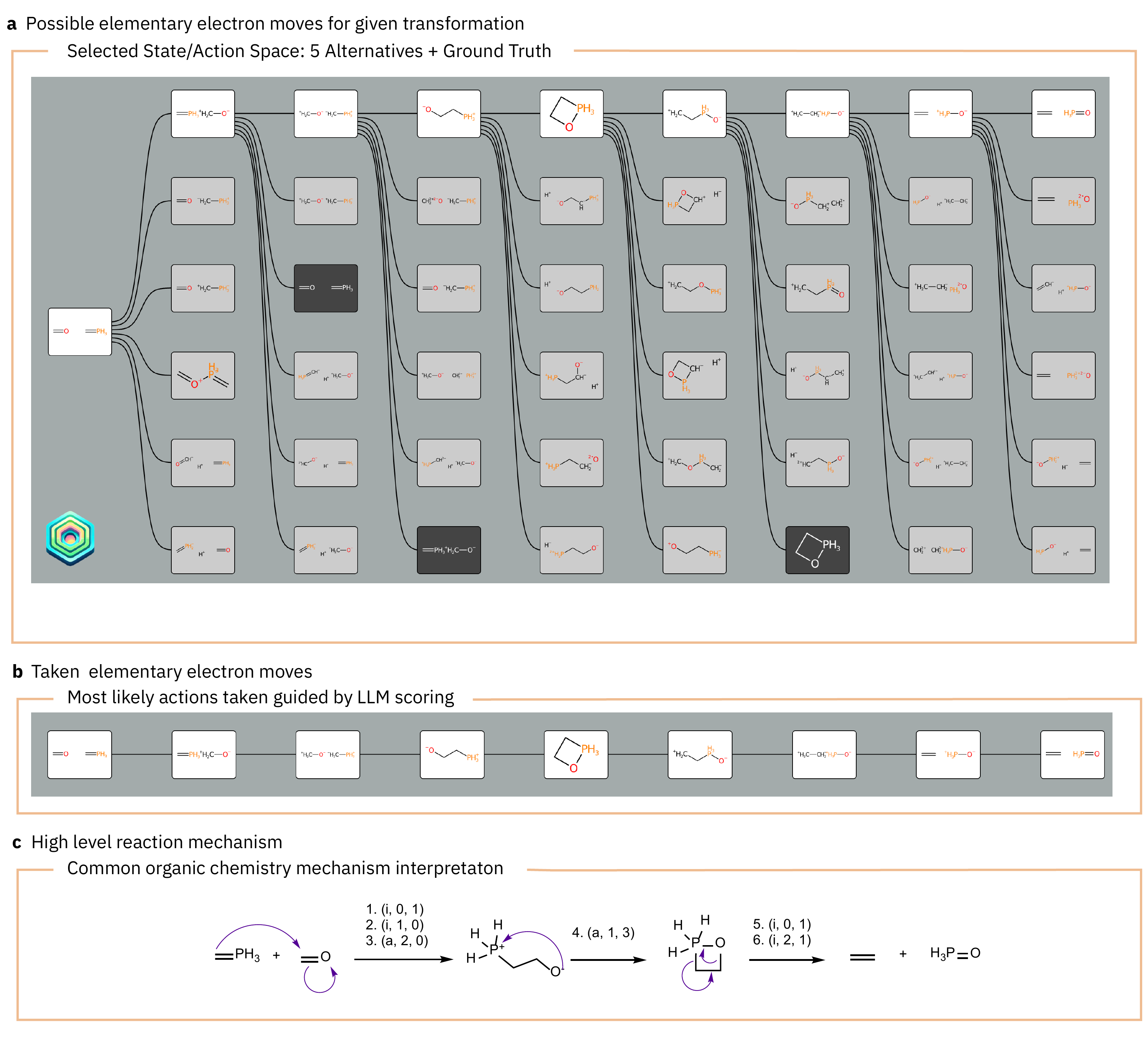}
    \caption{Description of Task 11 in the mechanistic benchmark. \textbf{a} shows the ground truth path (highlighted in white) along 5 other options for each step (light grey). Dark grey indicates moves that are part of the ground truth sequence but have already been traversed, serving as a check against loops in the prediction model. \textbf{b} Shows the isolated ground truth mechanism of the reaction.}
\end{figure}
\begin{figure}[ht!]
    \centering
    \includegraphics[width=\linewidth]{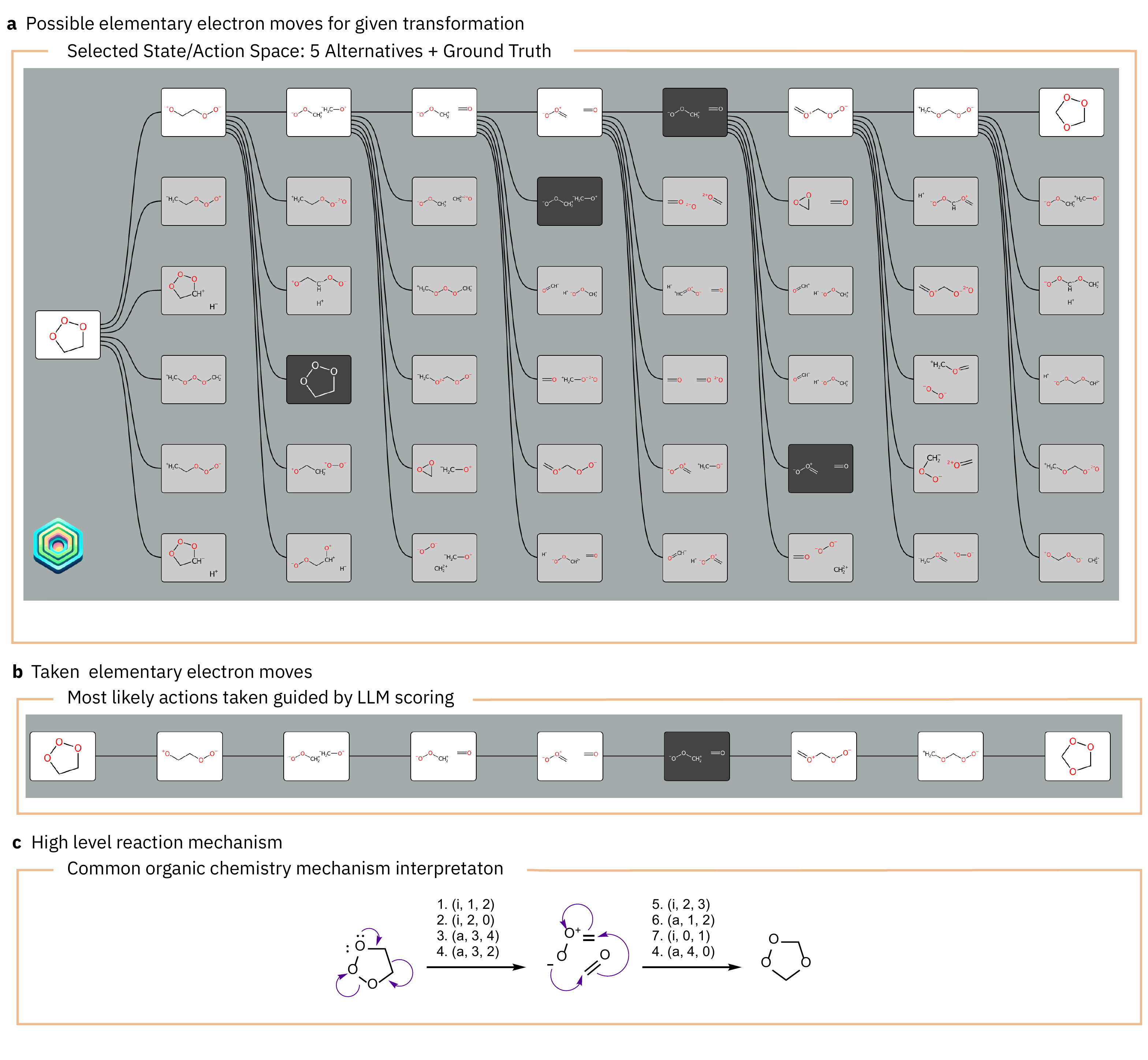}
    \caption{Description of Task 12 in the mechanistic benchmark. \textbf{a} shows the ground truth path (highlighted in white) along 5 other options for each step (light grey). Dark grey indicates moves that are part of the ground truth sequence but have already been traversed, serving as a check against loops in the prediction model. \textbf{b} Shows the isolated ground truth mechanism of the reaction.}
\end{figure}

\clearpage

\bibliographystyle{naturemag}
\bibliography{iclr2025_conference}